\documentclass{article}
\usepackage[T1]{fontenc}
\usepackage{iclr2027_conference,times}

\usepackage{amsmath,amssymb,amsthm}
\usepackage{booktabs}
\usepackage{graphicx}
\usepackage{subcaption}
\usepackage{microtype}
\usepackage{tabularx}
\usepackage{url}
\usepackage[hidelinks]{hyperref}

\newtheorem{proposition}{Proposition}
\newcommand{\Prob}{\mathbb{P}}
\newcommand{\ind}{\mathbf{1}}
\newcommand{\BinCDF}{F_{\mathrm{Bin}}}

\title{Available Guardrails: Certifying Selective Prediction across ML Systems}

\author{
Parivesh Priye$^*$ \\
Rivian and Volkswagen Group Technologies \\
\And
Yufeng Wang\thanks{These authors contributed equally to this work.} \\
Stony Brook University \\
\And
Haibin Ling \\
Westlake University \\
\And
Michael Chaykowsky \thanks{Corresponding author.} \\
Rivian and Volkswagen Group Technologies \\}

\iclrfinalcopy
\makeatletter
\g@addto@macro\@maketitle{\lhead{}}
\makeatother

\begin{document}
\maketitle

\begin{abstract}
A selective predictor acts as a safety gate: it returns an output only when the prediction appears sufficiently trustworthy. Deployments increasingly require this reliability to be \emph{certified} at a target precision for every reporting unit of interest, such as a tool, policy label, or patient subgroup. The main difficulty is often not whether a granted certificate is valid, but whether finite calibration data can produce one at all. As the gate becomes safer or more fine-grained, some units may receive too little evidence to certify. We make this notion of \emph{availability} computable through classical exact-binomial inversion and formulate reporting-partition selection, under a fixed group order, as a dynamic program that exposes the trade-off among safety, granularity, and served traffic. The resulting frontier reveals a large population opportunity that finite-sample estimation nearly erases: a truth-informed planner gains $0.157$ mean coverage over support balancing, whereas a naive estimator recovers only $0.005$, making recovery from finite data the central challenge. Constructing candidate partitions on one planning split and selecting among them on another recovers part of this gap, improving mean coverage over support balancing by $0.060$, with the direction reproduced in $59$ of $60$ model effects across three intent-routing datasets and two architectures. A complementary validity-preserving lever, reallocating the familywise error budget across reporting units, recovers additional coverage both with population quantities and noisy estimates. The same frontier recurs, with predictor-specific ceilings, across LLM tool-calling, content moderation, lesion classification, and recommendation. Certified availability is therefore a plannable deployment resource that determines when a safety gate can be certified, at what granularity, and over how much traffic.
\end{abstract}

\section{Introduction}

Modern AI systems are increasingly fronted by a gate. An agent decides whether to execute a tool call or defer to a human, a moderation system whether to auto-block a comment, a clinical model whether to commit to a diagnosis or route the case onward. Each is a \emph{selective predictor}: it acts only when its prediction appears sufficiently reliable and abstains otherwise \citep{chow1970,elyaniv2010,geifman2017}. Such a gate is trustworthy when the precision of the traffic it serves is \emph{certified} from data rather than assumed. A single global guarantee is often insufficient, because a gate that is safe on average may still be unsafe for a particular tool, policy category, or patient subgroup. Certifying each reporting unit separately exposes this hidden variation, but at a statistical cost: the same finite calibration set must support every unit's guarantee simultaneously.

That cost can eliminate an otherwise valid certificate entirely. Consider a target precision of $0.90$, a familywise failure probability of $0.05$, and a true selective error of $0.05$. Certifying a single group at $80\%$ availability requires $179$ served calibration examples; fifty groups require $450$ each for the same per-group availability. The error margin matters just as much: holding fifty groups fixed while raising the error from $0.05$ to $0.09$ increases the requirement from $450$ to $13{,}407$ examples. Reporting granularity is therefore constrained from two directions, by the number of simultaneous guarantees and by the margin each leaves between the true error and its target.

Finite-sample risk-control methods address a different part of the problem: they establish when a returned policy may be trusted \citep{bates2021,angelopoulos2025ltt,angelopoulos2024crc}, and some study the power to return a nonempty policy. They do not determine how a fixed label budget should be allocated among alternative reporting partitions at a requested granularity. That is our design question: once a designer may merge declared groups, which partition certifies the most traffic? At the population level it has an exact answer. Classical exact-binomial inversion \citep{clopper1934,brown2001,krishnamoorthy2007,thulin2014} gives the certificate availability of any frozen candidate in closed form; fixing the number of reporting units turns this into a per-partition score, and the best contiguous partition under a fixed reliability order follows from a dynamic program. Reaching that frontier from finite data is harder, because a planner that directly maximizes an estimated score is biased toward partitions whose estimation noise happens to be favorable. We reduce this coupling with a held-out split: one subset of the planning labels constructs a slate of candidate partitions, a disjoint subset selects among them, and certification sees only the chosen map, preserving the validity contract.

This paper makes four contributions. First, we identify certificate \emph{availability}, rather than validity alone, as a central obstacle to fine-grained selective prediction, and make it exactly computable from a candidate's support, error, safety target, and multiplicity. Second, we turn availability into a design objective, casting reporting-partition selection under a fixed reliability order as an exact dynamic program that traces the frontier among safety, granularity, and served traffic. Third, we develop a held-out selection procedure that constructs candidate partitions on one split of the planning data and selects among them on another, so the certification sample remains independent of the design choice and validity is preserved. Finally, we evaluate the resulting framework across synthetic generators, intent routing with two architectures, controlled distribution shift, and four additional gating domains spanning LLM tool-calling, content moderation, lesion classification, and recommendation, reporting both the regimes in which the approach helps and the boundaries at which it does not.

\section{Related Work}
\label{sec:related}

\textbf{Certifying validity from finite labels.}
The statistical toolkit for certifying validity from finite labels is classical and mature. Exact confidence intervals and binomial power analyses
\citep{clopper1934,brown2001,krishnamoorthy2007,thulin2014}
provide certificates for a single proportion, while Learn-Then-Test and conformal risk control extend the same finite-sample rigor to black-box selection and broader risk targets
\citep{bates2021,angelopoulos2025ltt,angelopoulos2024crc}.
A parallel line of work strengthens these guarantees under distribution shift. Covariate-shift conformal methods reweight examples to accommodate known changes in the input distribution
\citep{tibshirani2019}, LEC imposes a linear expectation constraint
\citep{wang2026lec}, SCoRE controls discovery risks through e-values
\citep{bai2026score}, and COIN constructs thresholds for selective question answering
\citep{wang2026coin}. This literature already goes beyond validity alone by studying power, namely whether finite data can return a useful, nonempty policy, and Learn-Then-Test in particular develops the split, fixed-sequence testing scheme that we reuse at certification time. However, each method operates under its own native certification contract, which differs from the groupwise contract introduced in Section~\ref{sec:method}. Rather than combine these methods into a single leaderboard, we compare numerical frontiers only under that shared contract and explicitly distinguish evaluations conducted under distribution shift. Our point of departure is therefore to treat validity as established and ask an earlier question: once reporting granularity becomes fine, how often can finite labels produce a certificate at all?

\textbf{Group-structured guarantees.}
A second line of work directly studies groups and the coverage costs induced by group-level guarantees, but addresses a different design problem concerning which partitions are admissible. Fair Risk Control and adaptive risk control extend finite-sample guarantees to richer structured group risks
\citep{zhang2024fair,blot2024}, while HG-CRC assumes a user-specified hierarchy and retreats from leaves to ancestors when a leaf cannot be certified, explicitly quantifying the participation cost of that retreat
\citep{salem2026}. In contrast, we compare flat partitions at a fixed requested number of reporting units in settings where the designer may merge declared groups. Our constrained experiment respects a declared hierarchy but is not an implementation of HG-CRC. The two perspectives are therefore complementary: HG-CRC adapts the reporting level within a supplied hierarchy, whereas our framework optimizes availability across admissible partitions at fixed granularity. Our later HWU experiment combines these views by requiring the availability optimizer to respect the semantic hierarchy.

\section{Methods}
\label{sec:method}

\subsection{Problem setting}

\begin{figure}[h]
\centering
\includegraphics[width=\textwidth]{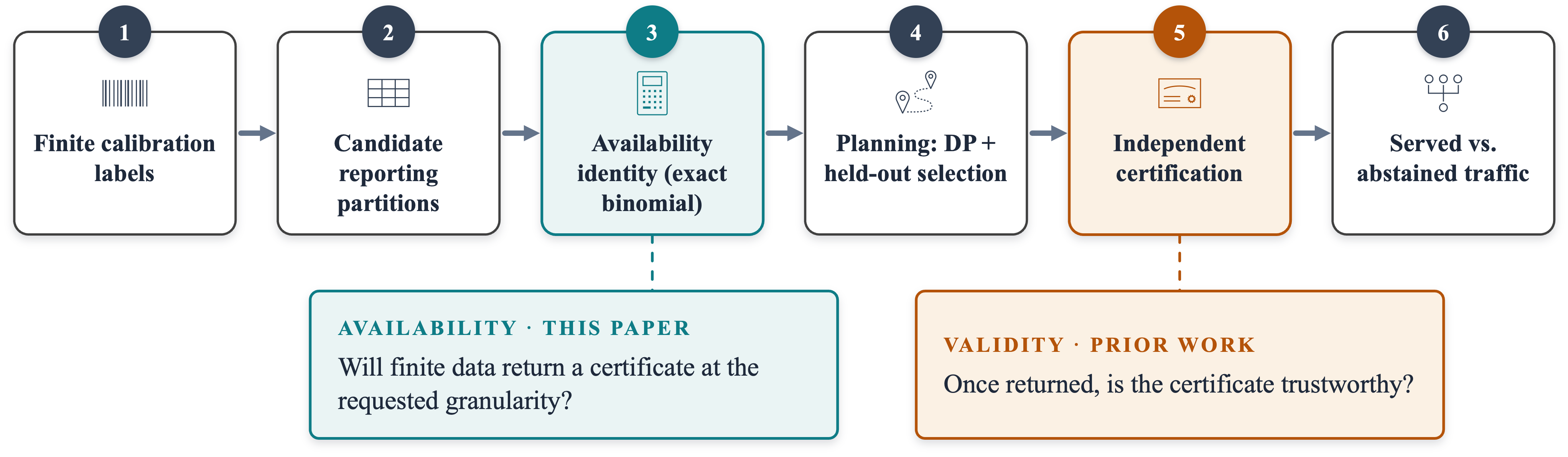}
\caption{Validity versus availability. Prior work asks whether a returned
certificate is trustworthy; we ask one step earlier whether finite calibration
data can return a certificate at the requested granularity at all.}
\label{fig:overview}
\end{figure}

We first formalize the certification contract, since the paper hinges on separating what a certificate guarantees from what finite data can actually deliver (Figure~\ref{fig:overview}). Let $(X,Y)\sim P$ and let a fixed predictor produce $\widehat Y=f(X)$, with correctness indicator $C=\ind\{\widehat Y=Y\}$. For other tasks, $C$ denotes the binary success event defined in Table~\ref{tab:contracts}. A frozen score $s(X)$ ranks examples by apparent reliability, with larger values indicating greater trust, and a threshold $t$ determines whether an example is served: $A_t=\ind\{s(X)\geq t\}$. Restricting to served examples, the selective error is

\begin{equation}
R(t)=\Prob(C=0\mid A_t=1),\qquad R(t)\leq\alpha=1-\tau.
\label{eq:risk}
\end{equation}

This gives a single global guarantee. The groupwise setting additionally introduces a mapping $G(X)\in\{1,\ldots,J\}$, fixed in advance using planning data, that assigns each served example to one of $J$ reporting units. Each unit $j$ has its own threshold $t_j$ and selective error $R_j(t_j)=\Prob(C=0\mid A_{t_j}=1,G=j)$. Deployment requires all certified units to satisfy their guarantees simultaneously:

\begin{equation}
\Prob_{\mathcal D_{\mathrm{cal}}}\!\left(
\forall j\text{ certified}:R_j(t_j)\leq\alpha\right)\geq1-\delta.
\label{eq:pac}
\end{equation}

The probability in Equation~\eqref{eq:pac} is over a fresh, independent and identically distributed (IID) certification sample. Four elements must be fixed before that sample is drawn: the score function, the reporting map, the within-unit ordering of candidate thresholds, and the certification distribution itself. Each later reappears as a potential boundary of the guarantee.

\subsection{From exact inversion to certificate availability}

All subsequent results rely on one exact identity, so we state it first. Consider a fixed candidate, meaning one reporting unit paired with one threshold, that serves $n>0$ certification examples and incurs $E$ errors. Define

\begin{equation}
\BinCDF(k;n,p)=\Prob\!\left(\mathrm{Bin}(n,p)\le k\right)
=\sum_{i=0}^{k}\binom{n}{i}p^{i}(1-p)^{n-i}
\label{eq:bincdf}
\end{equation}

as the binomial lower-tail probability of observing at most $k$ errors among $n$ independent trials with error probability $p$. The one-sided Clopper--Pearson upper confidence bound $U_{\mathrm{CP}}(E,n;\gamma)$ at failure level $\gamma$ \citep{clopper1934} is the largest error rate consistent with observing $E$ errors: when $E<n$, it is the unique $u\in(0,1)$ satisfying $\BinCDF(E;n,u)=\gamma$, and when $E=n$ it equals $1$. Certification succeeds exactly when this bound lies below the target, $U_{\mathrm{CP}}(E,n;\gamma)\leq\alpha$. The following equivalence converts that pass/fail event into a probability that can be used for planning.

\begin{proposition}[Exact availability identity]
For $0<\alpha,\gamma<1$,
\begin{equation}
U_{\mathrm{CP}}(E,n;\gamma)\leq\alpha
\quad\Longleftrightarrow\quad
\BinCDF(E;n,\alpha)\leq\gamma.
\label{eq:identity}
\end{equation}

Let $e^*(n,\alpha,\gamma)$ denote the largest integer satisfying the right-hand inequality, or $-1$ if no such integer exists. If the true selective error rate is $r$, then the candidate's certificate availability is

\begin{equation}
\pi(n,r;\alpha,\gamma)=\BinCDF(e^*;n,r).
\label{eq:availability}
\end{equation}

When no example is selected, set $\pi(0,r;\alpha,\gamma)=0$.
\end{proposition}

The identity follows from inversion of the exact binomial test. Its practical value is that substituting the candidate's \emph{true} selective error converts certification from an outcome observed after calibration into a probability predictable before it. For $J$ simultaneous units, Bonferroni correction sets $\gamma=\delta/J$, and two effects then govern the support required. Multiplicity is the milder one: for a fixed margin between $r$ and $\alpha$, the required support grows only on the order of $\log J$. The error margin matters far more, since as $r$ approaches $\alpha$ the availability in Equation~\eqref{eq:availability} falls sharply at any fixed budget. Any strictly positive margin remains certifiable in principle, but high availability can require rapidly increasing sample sizes.

\subsection{Population availability at a fixed granularity}

Different partitions create different reporting units, so comparing only the number of certified units is misleading: a method could improve that count simply by proposing coarser units. We therefore fix the requested granularity $J$ and evaluate all partitions at the same $J$, asking how much traffic they certify rather than how many units pass. For a segment $S$ served at threshold $t$, let $q_S(t)$ denote its traffic share, $r_S(t)$ its selective error, and $N$ the certification-sample size. We score a segment by its traffic multiplied by its probability of certification, evaluating that probability at the rounded expected selected count:

\begin{equation}
v(S,t;J)=q_S(t)\,
\pi\!\left(\lfloor Nq_S(t)\rfloor,r_S(t);
1-\tau,\delta/J\right).
\label{eq:segment}
\end{equation}

For a partition $\mathcal P_J$ containing exactly $J$ units, each segment retains its best threshold from a finite grid $\mathcal T$, giving the planning objective

\begin{equation}
U_J(\mathcal P_J)=\sum_{S\in\mathcal P_J}\max_{t\in\mathcal T}v(S,t;J).
\label{eq:utility}
\end{equation}

The deployed certifier does not commit to this single planned threshold. Within each unit it walks the threshold grid downward and retains the lowest threshold that passes before the first failure (Appendix~\ref{app:algorithm}). Two effects therefore separate $U_J$ from realized coverage, acting in opposite directions. Selected support is random rather than equal to its expectation, so substituting $\lfloor Nq_S(t)\rfloor$ into a nonlinear certification probability can be optimistic: in one small example a projected score of $0.500$ corresponds to exact expected coverage of only $0.2761$ (Appendix~\ref{app:exact-checks}). Working the other way, the downward walk can certify traffic beyond the planned threshold. Thus $U_J$ is an approximation, not a lower bound on deployed coverage.

The mean-support approximation can be replaced by exact integration over the random selected count (Appendix~\ref{app:exact-checks}), which is inexpensive at these problem sizes; we retain the projected objective as the default and measure the approximation error explicitly. The phrase ``exact dynamic program'' thus has a precise meaning: the recurrence exactly optimizes an approximate segment objective over the restricted family of contiguous cuts of a fixed ordering. Sweeping $J$ over a set $\mathcal J$ produces a frontier $\{(J,U_J):J\in\mathcal J\}$. Our primary quantity is \emph{traffic availability}, the traffic-weighted expected certified coverage, as distinct from \emph{unit availability}, whether a particular required unit certifies, and from the probability that all $J$ units certify at once. These differ when a rare unit matters more than its traffic share, as the HWU scenario-service analysis illustrates (Appendix~\ref{app:exact-checks}), and the application must ultimately decide which notion and which reporting resolution matter.

Optimizing over every partition of $K$ groups is intractable, so we restrict the search. We freeze a reliability order over the $K$ groups, sorting by planning error, then mean score, then identifier, and consider only contiguous cuts of it. Index the ordered groups by $0,\ldots,K-1$ and let $w_{ab}$ denote the best segment score $\max_{t\in\mathcal T}v(S_{ab},t;J)$ for the run spanning positions $a$ through $b$. The best score $D(j,b)$ for partitioning the first $b$ positions into $j$ runs then satisfies

\begin{equation}
D(j,b)=\max_{a\in\{j-1,\ldots,b-1\}}
\{D(j-1,a)+w_{a,b-1}\},\quad D(0,0)=0,
\label{eq:dp}
\end{equation}

so $D(J,K)$ returns the exact optimum over contiguous $J$-partitions in $O(K^2|\mathcal T|)$ time for segment construction and $O(JK^2)$ time for the recurrence. Semantic constraints are incorporated by setting $w_{ab}=-\infty$ whenever a proposed segment crosses a forbidden boundary. The same recurrence, supplied with different estimates of $q$ and $r$, defines the planners we compare: a nondeployable \emph{population oracle} that uses generator-level quantities, a \emph{direct plug-in planner} that estimates them from planning data, and three estimate-free baselines, namely support-balanced, equal-count, and random-order support balancing. The reliability ordering, tie-breaking rules, and baseline definitions are specified in Appendix~\ref{app:algorithm}.

\subsection{Held-out candidate selection}

Direct optimization over noisy estimates is optimistic, because the same noise that constructs candidate partitions also decides which one appears best. The held-out planner separates these roles. A construction subset produces eight candidate partitions: the direct plug-in, support-balanced, and equal-count partitions together with five random-order support-balanced candidates. A disjoint selection subset rescales each frozen candidate using Equation~\eqref{eq:segment} and keeps the highest-scoring one, ties resolved by a fixed preference order (Appendix~\ref{app:algorithm}). Only the selected map is retained, after which the score model and starting thresholds are refit on all planning labels. Splitting the planning data reduces the sample for both construction and selection, so held-out selection helps only when the reduction in selection bias outweighs the cost of smaller samples, a trade-off we measure empirically.

Certification remains independent of this entire planning procedure. Within each unit a fixed sequence walks downward from the planned threshold, tests each candidate at level $\delta/J$, and stops at the first failure \citep{angelopoulos2025ltt}; a union bound over the $J$ units then recovers Equation~\eqref{eq:pac} (Appendix~\ref{app:proofs}). Planning quality therefore changes which certificates are available and how much traffic they serve, while independent certification preserves the same validity guarantee.

\section{Results}
\label{sec:results}

We present the studies in the order that builds the argument, each addressing the
question the previous one raises, and we report where the planner loses as well as
where it wins. Unless a synthetic factor overrides them, the default precision target
is $\tau=0.90$ and the familywise level is $\delta=0.10$. The synthetic and intent
studies follow frozen, preregistered designs, while comparisons that reuse existing
real traces are exploratory. Each subsection introduces the datasets and outcomes it
uses, with complete protocols, dataset populations, and training recipes in
Appendices~\ref{app:protocols}, \ref{app:score}, and \ref{app:application-details}.

\subsection{Exact availability reveals a population opportunity}

\begin{figure}[h]
\centering
\begin{subfigure}[t]{0.48\textwidth}
\centering
\includegraphics[width=\textwidth]{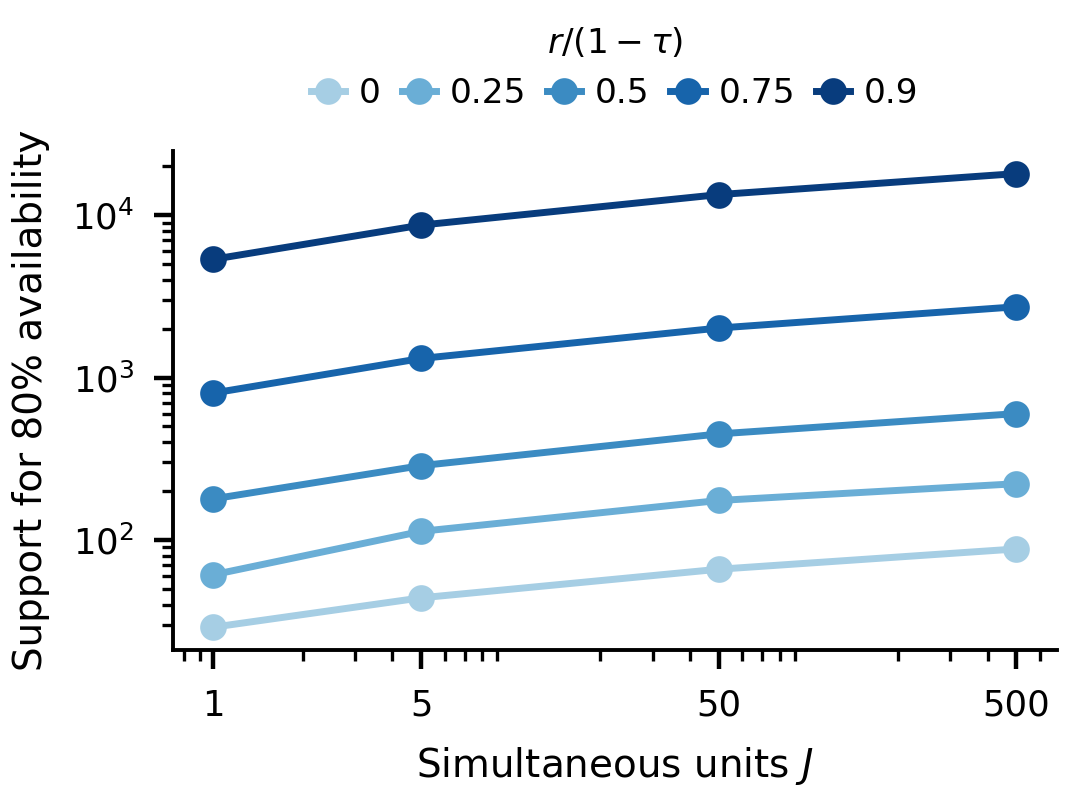}
\caption{Error margin dominates support.}
\label{fig:synthetic-a}
\end{subfigure}
\hfill
\begin{subfigure}[t]{0.48\textwidth}
\centering
\includegraphics[width=\textwidth]{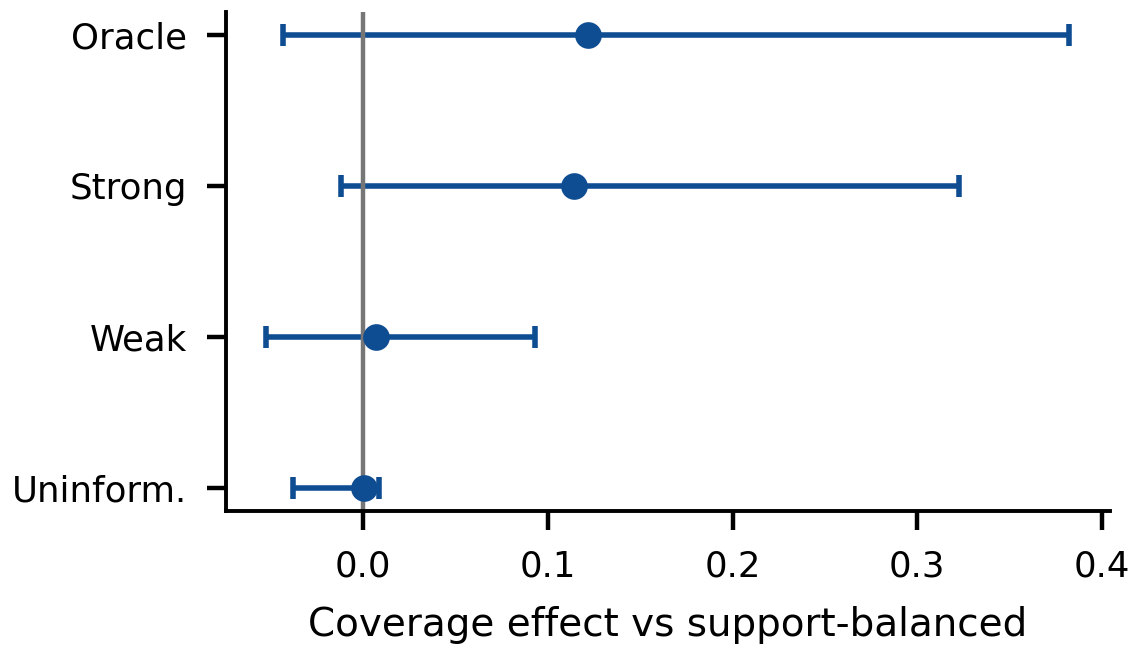}
\caption{Benefit requires score information.}
\label{fig:synthetic-b}
\end{subfigure}

\begin{subfigure}[t]{0.48\textwidth}
\centering
\includegraphics[width=\textwidth]{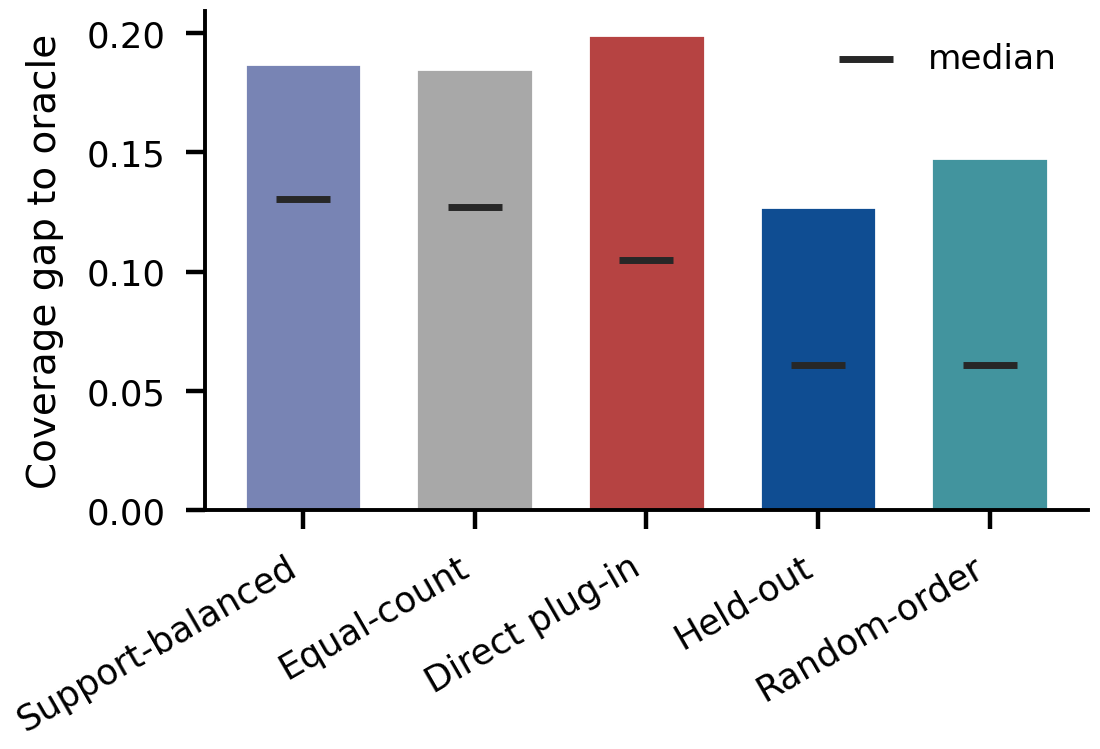}
\caption{Coverage gap to the population planner.}
\label{fig:synthetic-c}
\end{subfigure}
\hfill
\begin{subfigure}[t]{0.48\textwidth}
\centering
\includegraphics[width=\textwidth]{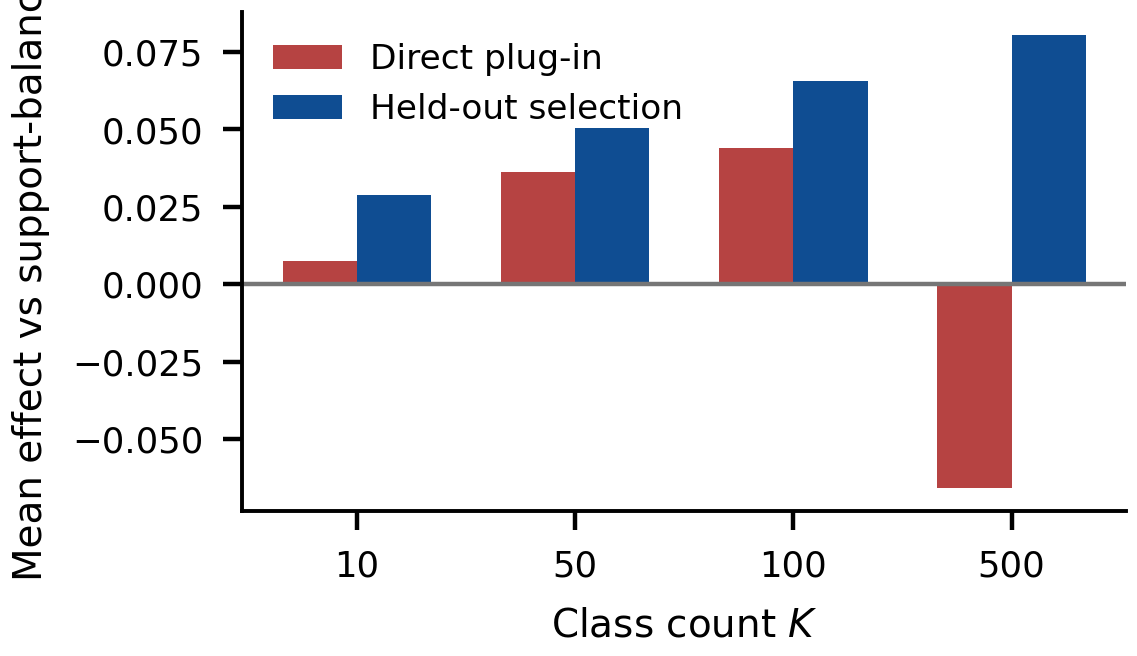}
\caption{Held-out selection at large $K$.}
\label{fig:synthetic-d}
\end{subfigure}
\caption{Synthetic evidence for the availability identity and the planners.}
\label{fig:synthetic}
\end{figure}

Before using the identity as a design objective, we first verify that it computes the claimed certification probability. Across $432$ untouched binomial cells, predicted certification frequency differs from simulated frequency by a mean absolute error of only $0.00067$. With the arithmetic verified, Figure~\ref{fig:synthetic}a shows the two forces that restrict granularity. Multiplicity is the milder one: holding true error at $0.05$ and availability at $80\%$, increasing $J$ from $1$ to $5$, $50$, and $500$ raises the required support only from $179$ to $287$, $450$, and $600$. The error margin is much more consequential. At fixed $J=50$, increasing the true error from $0.05$ to $0.09$ raises the same support requirement from $450$ to $13{,}407$.

The identity also shows that partitioning creates a substantial population opportunity. With true traffic and error rates, the population planner outperforms support balancing in $1{,}366$ of $1{,}845$ cell-by-$J$ configurations, ties in $446$, and loses in only $33$, for a mean coverage gain of $0.1568$. This is a genuine population effect rather than a numerical artifact. The difficulty appears when the planner must rely on finite estimates. Direct plug-in planning gains only $0.0050$ on average, has median improvement zero, and reaches $-0.3263$ at its 5th percentile. At $K=500$, where estimation is thinnest, its mean effect becomes negative at $-0.0659$. Several sources contribute to the oracle gap, including noisy segment and threshold estimates and error in the frozen reliability order, so the experiment does not isolate selection bias as the only cause. The availability identity therefore defines a useful objective, but finite estimation makes reaching its population frontier the real problem.

The objective itself is approximate. Exact analyses on small problems isolate three sources of discrepancy: replacing random support by its mean, planning around one starting threshold rather than the full threshold sequence, and restricting the search to contiguous cuts of a fixed order. Each can reduce coverage in individual cases, yet replacing the mean-support shortcut with exact support integration is not a consistently better planner in the tested design (Appendices~\ref{app:exact-checks} and \ref{app:planning-diagnostics}).

\subsection{Held-out selection recovers part of the opportunity}

Splitting the planning labels converts the near-miss above into a partial recovery. The held-out planner clears every preregistered gate for selection quality, regret to the oracle, and validity. Across $1{,}340$ non-endpoint configurations, it improves coverage over support balancing by $0.0601$ on average and $0.00146$ at the median. The largest gain occurs precisely where direct plug-in planning performs worst: at $K=500$, the mean effect becomes $+0.0806$. Separating the data used to construct a partition from the data used to select it reduces average overestimation in the matched-pool audit, although it does not eliminate it (Appendix~\ref{app:planning-diagnostics}). The method improves mean coverage over support balancing, but its $221$ losses in Table~\ref{tab:synth} show that it does not dominate configuration by configuration. The difference between mean and median also reveals that the gain is concentrated. Improvement exceeds $0.05$ coverage in roughly one third of configurations against support balancing but only one sixth against the stronger random-order baseline, and nearly half of configurations are ties or losses.

The oracle gap tells the same story. Support balancing leaves a median absolute coverage gap of $0.1305$, direct plug-in planning leaves $0.1051$, and held-out selection reduces it to $0.0611$. However, a plain random-order candidate reaches an almost identical gap of $0.0608$. Relative to that stronger baseline, the held-out advantage shrinks to a mean of $0.0208$ and a median of zero. Random-order support balancing is therefore a genuinely strong matched-unit comparator, and the practical gain of the planner should be judged against that baseline rather than support balancing alone.

Holding the candidate pool fixed separates two possible sources of improvement (Table~\ref{tab:mechanism}). Choosing on held-out data rather than reusing the construction data contributes $0.0480$ across $64$ synthetic conditions. By contrast, adding the structured candidates to five random groupings contributes only $0.0027$ with median zero. Most of the benefit therefore comes from independent selection rather than from the particular structured candidates in the pool (Appendix~\ref{app:exact-checks}).

\begin{table}[h]
\centering
\caption{Matched-pool mechanism study. Coverage differences are means over
64 fixed synthetic conditions, each averaging 20 independent plans and
100 nested certification draws. Starts remain frozen except in the last row.
Positive values favor the first method.}
\label{tab:mechanism}
\small
\begin{tabular}{@{}lrrr@{}}
\toprule
Comparison & Mean & Median & Win/tie/loss\\
\midrule
Held-out full minus construction selection & 0.0480 & 0.0032 & 53/0/11\\
Held-out full minus uniform full pool & 0.0318 & 0.0054 & 44/0/20\\
Held-out full minus empirical traffic & 0.0161 & 0.0002 & 38/1/25\\
Held-out full minus held-out random five & 0.0027 & 0.0000 & 29/4/31\\
Held-out random five minus random one & 0.0136 & 0.0012 & 44/0/20\\
Refitted full minus refitted random five & 0.0030 & $-0.00003$ & 24/3/37\\
\bottomrule
\end{tabular}
\end{table}

If structured partition search is not where most of the remaining opportunity lies, the familywise error budget is a more promising lever. The standard certifier divides the budget equally as $\delta/J$. A frozen reallocation $\gamma_1,\ldots,\gamma_J$ satisfying $\sum_j\gamma_j\le\delta$ preserves the same familywise validity guarantee while assigning more testing budget to units close to certification. At the population level, this increases coverage by $0.038$ on average and by as much as $0.115$ in the hardest regimes. The effect also survives estimation: a deployable allocator built from a single noisy planning sample gains $0.011$ mean coverage, is positive in $68\%$ of cases, and never exceeds the nominal familywise level, while an uncertainty-aware estimate improves further (Appendix~\ref{app:budget}). This gain is several times larger than the $0.0027$ contribution from structured candidate search, suggesting that budget allocation is a more promising route toward the population frontier.

Score quality determines whether there is an opportunity to recover at all. Figure~\ref{fig:synthetic}b compares four score regimes using the mean held-out advantage over support balancing, with bars spanning the cell-level 5th to 95th percentiles. With an oracle score, the held-out planner improves mean coverage by $0.1217$ and median coverage by $0.0882$; a strong learned score is nearly as effective, at $0.1144$ and $0.0851$. The effect collapses as ranking quality deteriorates: $0.0072$ mean and zero median for a weak score, and $0.0004$ mean and zero median for an uninformative one. Availability-aware planning therefore offers little when the score cannot distinguish correct from incorrect predictions. It becomes useful only when the underlying reliability ranking already contains substantial information.

\subsection{The IID frontier improves across two architectures}

\begin{figure}[h]
\centering
\begin{subfigure}[t]{0.48\textwidth}
\centering
\includegraphics[width=\textwidth]{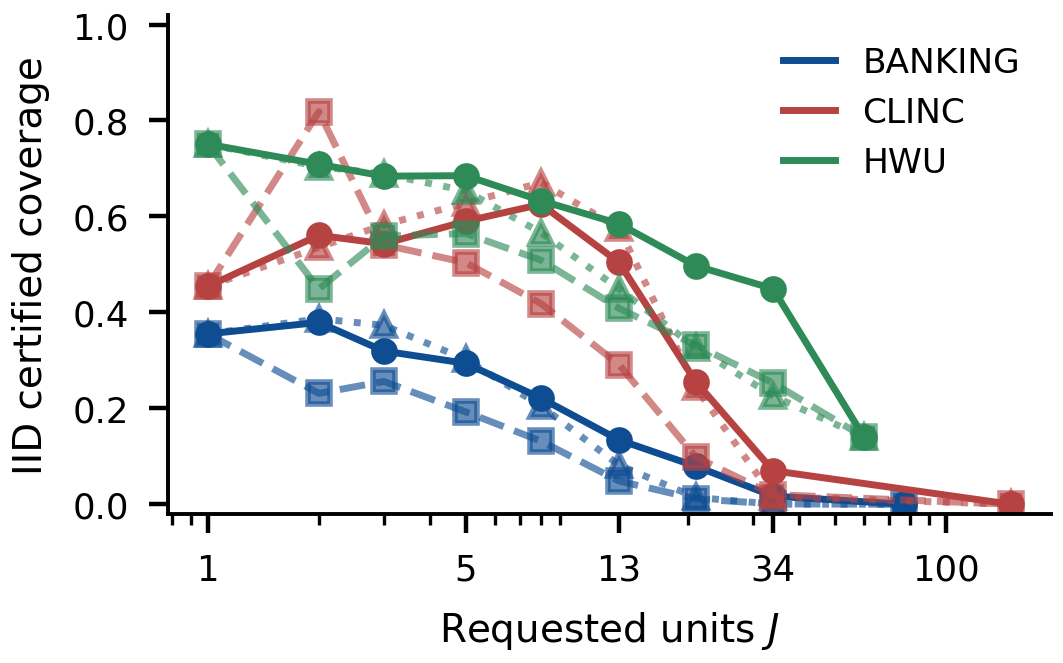}
\caption{DeBERTa: matched-unit frontiers.}
\label{fig:real-a}
\end{subfigure}
\hfill
\begin{subfigure}[t]{0.48\textwidth}
\centering
\includegraphics[width=\textwidth]{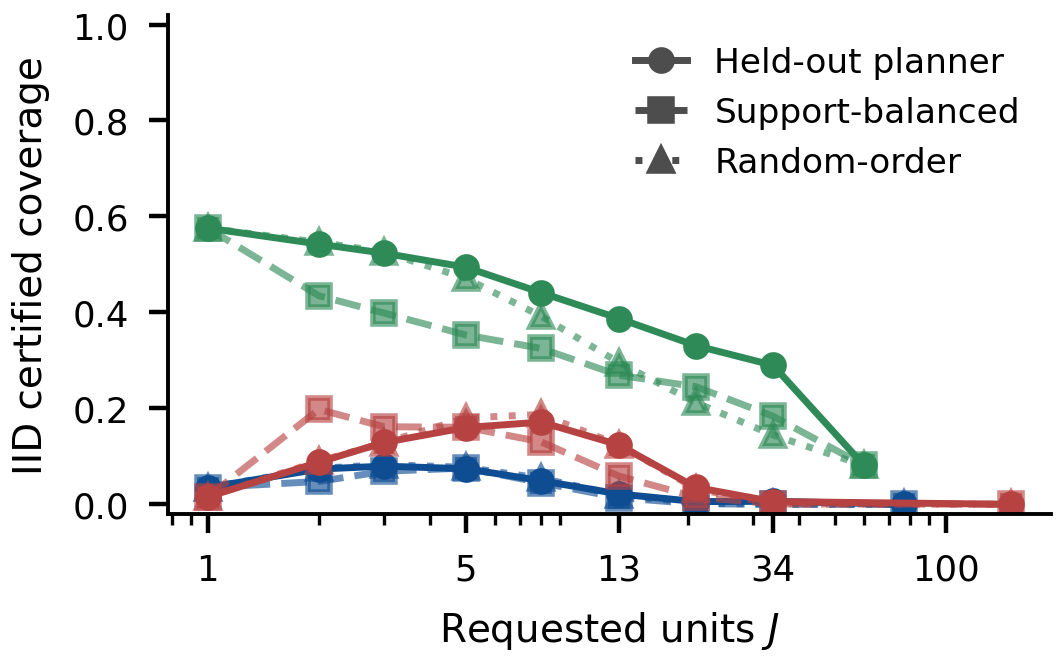}
\caption{DistilRoBERTa: matched-unit frontiers.}
\label{fig:real-b}
\end{subfigure}

\caption{Intent-classification certified-coverage frontiers, by architecture.}
\label{fig:real}
\end{figure}

The synthetic studies are controlled but artificial, so we next test whether the same pattern survives trained models. It does, at least in direction. For each architecture, Figure~\ref{fig:real} compares the held-out planner's IID certified-coverage frontier with support balancing and mean random-order support balancing, using one curve per dataset and averaging each point over thirty resplits within five independently trained models. Against support balancing, the held-out planner improves mean IID coverage at both prespecified frontier points in all six architecture-by-dataset strata (Figure~\ref{fig:real-checks}a). The gains range from $0.0045$ on DistilRoBERTa BANKING to $0.2128$ on DeBERTa CLINC. These sixty model-by-granularity effects come from thirty trained models, each evaluated at two granularities. All but one are positive, the exception being a single DistilRoBERTa BANKING effect at $J=13$ (Table~\ref{tab:real-effects}). The direction therefore replicates much more reliably than the magnitude, which varies by more than an order of magnitude across architectures and datasets.

Against the stronger random-order baseline, the result is intentionally less uniform. At $J=13$, the held-out planner exceeds mean random-order coverage on DeBERTa BANKING and HWU but trails it on CLINC. DistilRoBERTa differences are smaller and straddle zero, while direct plug-in planning retains higher coverage in several BANKING and HWU settings. These outcomes are consistent with the synthetic mechanism study: the gain from independent selection is robust, but the structured candidate family itself does not dominate a strong random baseline.

Selection frequencies show how the planner responds to this heterogeneity, although they do not by themselves explain performance. Across $16{,}200$ real selection decisions, support balancing is chosen $28.2\%$ of the time, direct plug-in $17.3\%$, equal-count $5.4\%$, and random-order partitions $49.0\%$. No candidate family dominates. The planner instead adapts its choice to the observed stratum. The earlier $0.0208$ margin over random-order balancing is a synthetic result and should not be interpreted as a summary of these real-data selection frequencies.

\subsection{Semantic constraints reduce the attainable frontier}
\label{sec:semantic}

A statistically favorable partition can still be operationally meaningless because the availability objective does not know what a reporting group represents. The HWU hierarchy allows us to quantify this distinction. Forbidding merges across scenario boundaries, which preserves semantic interpretability, reduces IID coverage for DeBERTa by $0.246$ and $0.167$ at $J=21$ and $J=34$, and for DistilRoBERTa by $0.229$ and $0.225$ (Figure~\ref{fig:real-checks}b). Every model-level effect is negative. The reliability-ordered frontier is therefore not a proved ceiling on the semantically constrained problem; it simply searches a larger set of merges under a different group ordering. In applications with named, fairness-sensitive, or legally defined groups, those boundaries must remain fixed or the admissible hierarchy of merges must be specified in advance.

Under this constraint, the held-out planner is no longer the strongest procedure in the HWU replay. Holding the scenario restriction fixed for every candidate across ten resplits of each of ten HWU models, the direct constrained planner outperforms the full held-out planner in all twenty model-by-granularity comparisons at $J=21$ and $34$. Mean margins range from $0.0267$ to $0.0518$ across the four architecture-by-granularity settings (Table~\ref{tab:constrained}). Because the direct planner also optimizes how reporting units are allocated across scenarios, the comparison is between two complete procedures rather than a clean isolation of the selection step. It also reaches $0.06$ to $0.20$ more named scenarios on average, counting a scenario whenever at least one of its units certifies. Even so, neither method reaches six of the eighteen scenarios on average, leaving some intents entirely unserved (Table~\ref{tab:scenario-service}).

\subsection{Certificate transfer depends on the form of shift}

All guarantees so far assume that certification and deployment share the same distribution. This assumption can fail, and every formal guarantee in this paper applies to the certification distribution only. The issue is concrete in CLINC: the official test mixture contains $18.18\%$ out-of-scope traffic, compared with $6.25\%$ in the evaluation pool, so we treat it throughout as a shifted deployment distribution. The effect of this mismatch depends on which component of the distribution changes.

A label-prior shift can be repaired, but only by paying in coverage. At $30\%$ out-of-scope prevalence, a stale policy's precision falls to $0.753$. Exact reweighting restores precision to $0.935$ while reducing coverage to $0.455$, whereas estimated reweighting preserves more coverage at lower precision. Conditional-outcome shift is more severe. When $P(X)$ is unchanged, feature-density ratios contain no corrective signal. At $5\%$ shift severity, fresh labels restore precision to $0.913$ only by roughly halving coverage, and at $10\%$ severity recalibration abstains entirely. Monitoring also reacts too slowly to protect already served traffic: within $2{,}000$ probes, it detects only $11.3\%$ of a $2.5\%$ conditional shift (Appendix~\ref{app:shift}).

\subsection{The safety-availability frontier across four domains}

Intent routing represents only one task family, so we finally ask whether the same trade-off appears in qualitatively different safety gates. We hold the method fixed and evaluate four additional domains: a Qwen2.5-14B agent making single-turn tool calls on the Berkeley Function-Calling Leaderboard (BFCL), a Civil Comments moderator deciding which comments to auto-block, a fine-tuned ResNet-18 classifying skin lesions on DermaMNIST, and a matrix-factorization recommender predicting item preference with genre as the reporting group. Each domain uses one frozen base predictor, so repeated splits measure allocation variability rather than variability across independently trained models. These experiments therefore characterize the frontier for the specific predictor and candidate population rather than comprehensive deployment safety.

Each domain exhibits its own finite-budget coverage profile (Table~\ref{tab:md-curves}), with a ceiling determined by predictor quality as well as the certification procedure. The tool-call gate remains fully available at a $0.70$ target but is nearly closed at $0.90$, whereas the stronger moderation classifier retains high coverage even at $0.90$. Increasing the safety target reduces availability in every domain, but increasing granularity need not reduce it monotonically. Separate per-unit thresholds can initially offset the loss of support, as in recommendation, where coverage increases from $0.65$ at the global gate to $0.79$ at $J=2$ for $\tau=0.80$, before support starvation eventually dominates. The held-out planner is again mixed against stronger controls. Relative to support balancing, it gains $0.176$ on moderation and $0.051$ on function calling at their chosen operating points. On a shared candidate pool with refitting, however, it trails simple random-only selection on both tasks (Appendix~\ref{app:multidomain-curves}). The apparent advantage therefore depends on both the comparator and the selection protocol, reinforcing the same conclusion as the synthetic and intent studies: availability is a real deployment frontier, but no single planner dominates across all regimes.

\section{Conclusion and Limitations}

In this paper, we separate two failure modes traditionally conflated within a single certificate: validity, which bounds the error of an established policy, and availability, which determines whether finite data can produce a passing policy at a requested granularity. By leveraging exact-binomial power to quantify this constraint, safety gates can be planned against three coupled quantities: the safety of the served guarantee, its reporting granularity, and the necessary traffic rejection rate. This trade-off recurs with predictor-specific ceilings across intent routing, tool-calling, moderation, lesion classification, and recommendation, establishing certified availability as a fundamental deployment resource rather than a dataset artifact. The deeper statistical lesson is that while optimal partitioning offers substantial theoretical value, empirical optimizers largely lose this advantage to estimation noise. Although held-out selection recovers a meaningful share of this potential through candidate diversity rather than structured search, larger gains lie in reallocating the family-wise error budget while preserving validity. Consequently, making a guarantee available from finite data emerges as a significant statistical problem in its own right.

These conclusions are subject to important boundaries. First, while the benefits of held-out selection are qualitatively reliable, their magnitude varies by orders of magnitude and diminishes with uninformative scores. Second, the planner is limited by a weak reliability prior, depending primarily on candidate selection rather than ordering accuracy. Finally, these guarantees apply strictly to their explicit contracts: they do not accommodate adaptively chosen candidates or unmitigated distributional drift, which requires reweighting or fresh calibration data. Within a fixed contract, informative scoring, and defined granularity, certified availability remains a robust tool for deployment planning.

\label{end:main}
\section*{Ethics Statement}

This work uses only public benchmarks, with no new data collection or human
subjects, and measures benchmark success rather than clinical efficacy or
deployment safety. Its two principal ethical stakes are addressed in the body of
the paper. First, merging groups by statistical reliability can fuse categories
that a deployment must keep distinct, so named, fairness-sensitive, or legally
defined groups should be preserved or governed by a declared hierarchy
(Section~\ref{sec:semantic}). Second, a certificate can appear valid after the
deployment distribution has shifted, which our controlled-shift study is
designed to expose rather than hide.

\section*{Reproducibility Statement}

The availability identity is proved in Appendix~\ref{app:proofs} and the full
planner in Appendix~\ref{app:algorithm}, while Section~\ref{sec:results}
separates the frozen studies from the exploratory ones. Under a pinned
environment and fixed seeds the pipeline is deterministic, and the supplementary
material contains the code, aggregate and seed-level results, and an automated
check that recomputes every reported number from them
(Appendix~\ref{app:release}).

\section*{AI Statement}

A large language model assisted the authors in polishing the language of the manuscript. The authors verified that all claims, proofs, mathematical formulations, and reported values are valid, checked them against the implementation and results, and take full responsibility for the manuscript.

\bibliography{references}
\bibliographystyle{iclr2027_conference}

\appendix

\section*{Appendix overview}

The appendices are organized by purpose. We begin with methodology and
guarantees: the proofs behind the availability identity, the dynamic program,
and the validity of the certified policy, followed by the full planner
specification, the study protocols and reporting conventions, and the score
model and training recipe. We then present the experimental evidence, one
section per scientific question: extended synthetic results, independent-model
comparisons, controlled distribution shift, cross-domain frontiers,
exact-coverage and additional-planner analyses, and a study of familywise-budget
allocation. The final two sections give the application populations and the
reproducibility and artifact-integrity statement.

\section{Proofs}
\label{app:proofs}

Throughout, $\BinCDF(k;n,p)=\Prob(\mathrm{Bin}(n,p)\le k)$ is the binomial
lower-tail probability defined in Equation~\eqref{eq:bincdf}, and every
candidate, map, score, and threshold is frozen before the certification sample
is drawn.

\subsection{Proof of the exact availability identity}
We establish the two displayed claims in turn.

\emph{Step 1: the inversion identity~\eqref{eq:identity}.}
Fix $0\le E<n$ and regard the lower tail as a function of the error rate:
\begin{equation*}
g(p)=\BinCDF(E;n,p)=\sum_{i=0}^{E}\binom{n}{i}p^{i}(1-p)^{n-i}.
\end{equation*}
Differentiating yields a telescoping cancellation, leaving
\begin{equation*}
g'(p)=-\,n\binom{n-1}{E}p^{E}(1-p)^{\,n-1-E}<0
\qquad(0<p<1),
\end{equation*}
so $g$ is continuous and strictly decreasing on $[0,1]$, with $g(0)=1$ and
$g(1)=0$. For each $\gamma\in(0,1)$ there is therefore a unique $u\in(0,1)$
with $g(u)=\gamma$, and by definition
$u=U_{\mathrm{CP}}(E,n;\gamma)$. Because $g$ is strictly decreasing, the
inequality $u\le\alpha$ reverses under $g$:
\begin{equation*}
U_{\mathrm{CP}}(E,n;\gamma)\le\alpha
\;\Longleftrightarrow\;
g(u)\ge g(\alpha)
\;\Longleftrightarrow\;
\BinCDF(E;n,\alpha)\le\gamma,
\end{equation*}
which is Equation~\eqref{eq:identity}. When $E=n$, we have $g\equiv1$, so
$U_{\mathrm{CP}}=1>\alpha$ and $\BinCDF(n;n,\alpha)=1>\gamma$; both sides are
false and the equivalence holds vacuously.

\emph{Step 2: the availability formula~\eqref{eq:availability}.}
For fixed $n,\alpha,\gamma$, the map $k\mapsto\BinCDF(k;n,\alpha)$ is
nondecreasing, so the acceptance set $\{k:\BinCDF(k;n,\alpha)\le\gamma\}$ is a
(possibly empty) prefix $\{0,1,\ldots,e^*\}$ where
\begin{equation*}
e^*=e^*(n,\alpha,\gamma)=\max\{k\in\{0,\ldots,n\}:\BinCDF(k;n,\alpha)\le\gamma\},
\end{equation*}
and $e^*=-1$ precisely when even $k=0$ fails, i.e.\ when
$(1-\alpha)^{n}>\gamma$. By Step~1 the candidate certifies if and only if
$\BinCDF(E;n,\alpha)\le\gamma$, equivalently $E\le e^*$. Under the true
selective error $r$ the error count satisfies $E\sim\mathrm{Bin}(n,r)$, so the
probability of certifying is
\begin{equation*}
\pi(n,r;\alpha,\gamma)=\Prob(E\le e^*)=\BinCDF(e^*;n,r),
\end{equation*}
with the convention $\BinCDF(-1;n,r)=0$. This is
Equation~\eqref{eq:availability} and its boundary case $\pi=0$ when no error
count certifies.
\hfill$\square$

\subsection{Optimality of the dynamic program}
Fix the granularity $J$ and the reliability order, and index its groups by
positions $0,\ldots,K-1$. For $0\le a\le c\le K-1$, let
$w_{ac}=\max_{t\in\mathcal T}v(S_{ac},t;J)$ denote the best score of the run
spanning positions $a$ through $c$, and let $D(j,b)$ be the maximum of $\sum w$
over all partitions of the first $b$ positions $\{0,\ldots,b-1\}$ into exactly
$j$ contiguous runs, with $D(0,0)=0$ and $D(j,b)=-\infty$ when no such
partition exists (for instance when $j>b$). The
objective~\eqref{eq:utility} is a sum of per-run terms, and each run maximizes
its own threshold independently, so the score of a partition equals the sum of
its runs' $w$ values; the problem therefore has optimal substructure.

Any contiguous partition of $\{0,\ldots,b-1\}$ into $j\ge1$ runs has a last run
$\{a,\ldots,b-1\}$ for a unique split point $a$. Requiring every run to be
nonempty forces $a\in\{j-1,\ldots,b-1\}$. Removing the last run leaves a
contiguous partition of $\{0,\ldots,a-1\}$ into $j-1$ runs, whose score is at
most $D(j-1,a)$ and equals it for an optimal remainder. Adding the last run's
score $w_{a,b-1}$ and maximizing over $a$ yields
\begin{equation*}
D(j,b)=\max_{a\in\{j-1,\ldots,b-1\}}\bigl\{D(j-1,a)+w_{a,b-1}\bigr\},
\end{equation*}
which is Equation~\eqref{eq:dp}. Induction on $j$ and then $b$ shows that
$D(J,K)$ is the optimal score over all contiguous $J$-partitions of the order,
and the split points attaining each maximum reconstruct the optimal reporting
map. Setting $w_{ac}=-\infty$ whenever the run $S_{ac}$ crosses a forbidden
semantic boundary removes exactly the infeasible partitions and leaves the
argument otherwise intact. Non-contiguous partitions of the order lie outside
the optimization domain by construction.
\hfill$\square$

\subsection{Validity of the certified policy}
Let $\mathcal D_{\mathrm{plan}}$ denote the planning labels and
$\mathcal D_{\mathrm{cal}}$ the fresh, independent certification sample of
Equation~\eqref{eq:pac}. Condition on $\mathcal D_{\mathrm{plan}}$: the score
$s$, the reporting map $G$, the unit count $J$, and each unit's ordered list of
candidate thresholds are then deterministic functions of
$\mathcal D_{\mathrm{plan}}$ and do not depend on $\mathcal D_{\mathrm{cal}}$.
Fix a unit $j$ and test its thresholds in the pre-specified order, each by the
exact test of Proposition~1 at level $\delta/J$, stopping at the first that
fails and certifying the last that passed. Declaring unit $j$ safe with a
threshold whose true selective error exceeds $\alpha$ requires rejecting the
first true null in this ordered family, and the fixed-sequence testing guarantee
\citep{angelopoulos2025ltt} bounds that event's probability by the per-test
level:
\begin{equation*}
\Prob\!\left(j\text{ certified and }R_j(t_j)>\alpha
\;\middle|\;\mathcal D_{\mathrm{plan}}\right)\le\frac{\delta}{J}.
\end{equation*}
A union bound over the $J$ units then gives
\begin{equation*}
\Prob\!\left(\exists\,j\text{ certified}:R_j(t_j)>\alpha
\;\middle|\;\mathcal D_{\mathrm{plan}}\right)
\le\sum_{j=1}^{J}\frac{\delta}{J}=\delta,
\end{equation*}
and taking expectation over $\mathcal D_{\mathrm{plan}}$ removes the
conditioning while preserving the bound, which is exactly
Equation~\eqref{eq:pac}. The map, score, and threshold order enter only through
the conditioning, so candidate quality and selection bias affect which policy is
certified but never the level at which it is certified.
\hfill$\square$

\section{Planner Specification}
\label{app:algorithm}

\subsection{Inputs}
The planner receives a frozen predictor, declared groups, target $\tau$,
familywise failure probability $\delta$, requested granularity $J$,
certification size $N$, and grid $(0,0.02,\ldots,1)$. Only planning labels are
available at this stage.

\subsection{Reliability order}
For each predicted class, compute its construction error rate. Sort classes by
ascending error, descending mean score, and ascending integer class index.
Empty classes receive error one and mean score zero. The equal-count partition
assigns rank $k$ to $\min(J-1,\lfloor kJ/K\rfloor)$. The support-balanced
partition assigns each ordered class to
$\min(J-1,\lfloor Jc/S\rfloor)$, where $c$ is its preceding cumulative support
and $S$ is total construction support. If fewer than $J$ units are nonempty,
use the equal-count partition. Random-order support-balanced partitions apply
the same rule after seeded permutations.

\subsection{Segment tables and tie rules}
For every contiguous segment and grid threshold, compute the selected
construction count $m$, its error count $e$, and the estimated traffic share
$\widehat q=m/n_{\mathrm{plan}}$, where $n_{\mathrm{plan}}$ is the size of the
construction set. The projected certification support is
$\lfloor N\widehat q\rfloor$, and the estimated error rate is the Jeffreys mean
$\widehat r=(e+1/2)/(m+1)$. Score the pair with
Equation~\eqref{eq:segment}. Within a segment, maximize this score, then
projected support, and finally prefer the lower threshold. The dynamic program
uses the first maximizing split. Semantic constraints render cross-boundary
segments infeasible.

\subsection{Held-out candidate-selection algorithm}
Deterministically permute the planning role and divide it into equal
construction and selection subsets. Fit the correctness score on construction
labels. Build the direct plug-in, support-balanced, equal-count, and five
random-order support-balanced partitions. Freeze their maps and construction
thresholds.

On the selection subset, recompute $q$ and the Jeffreys estimate of $r$ for
each candidate at its fixed construction thresholds. Select the candidate
with the largest summed projected coverage from Equation~\eqref{eq:segment},
without a new threshold search. Ties within $10^{-12}$ prefer support-balanced,
equal-count, direct plug-in, then
random-order candidates by index. Retain the selected map, refit the score on
all planning labels, and update its starting thresholds. On certification data,
test grid points downward from each planned start and stop at the first failure.
A reporting unit that never passes abstains. Coverage and feasibility summaries
retain all abstain-all outcomes.

\section{Study Protocols and Reporting}
\label{app:protocols}

This appendix collects the full protocol behind each study summarized in
Section~\ref{sec:results}: sample sizes, resplit schemes, generator factors,
and the reporting conventions dictated by each study's independent unit.

\subsection{Synthetic}
An analytic check draws 432 hash-selected binomial cells (10{,}000 simulations
each) to validate the identity itself. The planner studies run on a generator
that crosses six label budgets, four class counts $K\in\{10,50,100,500\}$,
balanced or Zipf traffic, error heterogeneity, three targets, three failure
levels, and four score regimes ranging from oracle to uninformative. Two
disjoint confirmation pools carry the claims: 222 untouched cells for the
direct planner and 215 for the held-out planner, each with 25 planning draws
and 20 nested certification draws. The held-out aggregate excludes the
coincident endpoints $J=1,K$. Because the two pools differ, we do not divide
their mean gains to claim a recovered fraction. A further shared-candidate
study uses 64 conditions, 20 planning draws per setting, and 100 nested
certification draws to separate the benefit of held-out selection from that of
candidate diversity (Appendix~\ref{app:exact-checks}). Real-data planners
search the threshold grid $\mathcal T=\{0,0.02,\ldots,1\}$, while the synthetic
score generator uses $\{0,0.05,\ldots,0.95\}$.

\subsection{Intent classification}
On BANKING77 \citep{casanueva2020banking}, CLINC150 \citep{larson2019clinc}, and
HWU64 \citep{liu2019hwu}, we train five DeBERTa-v3-base
\citep{he2021debertav3} and five DistilRoBERTa-base \citep{liu2019roberta}
models each, analyzing the two architectures separately. Within each model, 30
deterministic resplits divide the evaluation mixture into 10/10/60/20\% for
construction, selection, certification, and IID deployment. The official test
trace is reported as a shifted mixture and held out of the IID analysis. The
model seed, an independent training run from a fresh random initialization,
serves as the outer independent unit, with resplits treated as nested
measurements. HWU's 18 scenarios supply the semantic hierarchy, and controlled
CLINC shifts (Appendix~\ref{app:shift}) separate label-prior from
conditional-outcome drift.

\subsection{Cross-domain safety gates}
Four further domains each gate a single frozen predictor: a Qwen2.5-14B agent
on BFCL single-turn function calling (four tool categories, official AST
checker), a logistic-regression moderator on Civil Comments, a fine-tuned
ResNet-18 on DermaMNIST (seven lesion classes), and a matrix-factorization
recommender on MovieLens (item liking, grouped by genre). Each domain uses one
frozen base predictor, with auxiliary models contributing score features. We
average ten resplits at each of $\tau\in\{0.70,0.80,0.90\}$, capping evaluation
pools at 8{,}000 rows. The separate held-out comparisons use six resplits at
domain-specific targets. These measure allocation variability within fixed
prediction records, not independent training replication.
Appendix~\ref{app:application-details} gives the populations and the additional
ten-resplit shared-candidate comparison.

\subsection{Reporting}
Because the independent unit differs across studies, so does the reporting
format. Synthetic analyses report cell-by-$J$ paired effects, win/tie/loss
counts, quantiles, and absolute coverage gaps to the population planner.
Planning draws are the independent simulation units, with certification draws
nested within plans. These summaries describe the fixed design. Intent analyses
instead expose every model-level effect individually, along with sign counts and
leave-one-model-out ranges. With five models per cell, the smallest two-sided
exact sign $p$-value is 0.0625, so we report the direction and magnitude of each
effect without making threshold-based significance claims.

\section{Score Model and Training}
\label{app:score}

The correctness score is deliberately simple and deterministic: a logistic
regression (at most 2,000 iterations, random seed zero) fitted on confidence,
top-two margin, entropy, Monte Carlo disagreement, and auxiliary-model agreement
and confidence, augmented with planning class accuracy and log support as
group-level features. When construction labels are constant, the model falls
back to a constant probability, with raw confidence as a secondary score. In
each cross-domain gate the same score is refit from whatever per-example
confidence signals the predictor exposes.

The intent classifiers set the training recipe. Both architectures train for
four epochs at a maximum sequence length of 64 with AdamW ($2\times10^{-5}$,
weight decay 0.01, 6\% warmup), using batch size 64 for DeBERTa and 128 for
DistilRoBERTa. HWU's 18 source scenarios supply the semantic hierarchy imposed
later as a constraint. DistilRoBERTa in-scope test accuracies across five seeds
were 0.6481 to 0.6932 on BANKING, 0.8533 to 0.8656 on CLINC, and 0.7471 to
0.7581 on HWU. These models broaden the evaluated performance and score-quality
range but are not matched-accuracy replacements for DeBERTa.

\section{Extended Synthetic Results}

The direct and held-out confirmations drew on disjoint pools of cells,
partitioned by a hash of each cell's identifier under their respective frozen
designs. The direct study contained 1{,}845 cell-by-$J$ configurations.
Relative to support balancing, direct plug-in planning won in 581
configurations, tied in 832, and lost in 432. Its mean effect was 0.00496, with
5th-to-95th percentiles spanning $-0.3263$ to 0.2855. When broken down by
condition, the mean effects were $-0.0659$ at $K=500$, $-0.0229$ with weak
scores, and $-0.0085$ under Zipf traffic.

The held-out study contained 1{,}340 nonendpoint configurations.
Table~\ref{tab:synth} collects their primary aggregate comparisons. A separate
validity audit checks the implementation against the proved contract: across
1{,}770 cell-by-$J$ rows it detected no excess after Bonferroni correction,
with the smallest excess-test $p$-value being 0.5604.

\begin{table}[h]
\centering
\caption{Held-out planner coverage effects at matched $J$. Each effect is a
cell-level mean over independent planning draws; certification draws are
nested within plans.}
\label{tab:synth}
\small
\begin{tabular}{@{}lrrrrr@{}}
\toprule
Baseline & Mean & Median & Wins & Ties & Losses \\
\midrule
Support-balanced & 0.06007 & 0.001465 & 744 & 375 & 221 \\
Equal-count & 0.05811 & 0.000857 & 723 & 375 & 242 \\
Direct plug-in & 0.07213 & 0.000009 & 673 & 381 & 286 \\
Mean random-order & 0.02079 & 0.000000 & 548 & 359 & 433 \\
\bottomrule
\end{tabular}
\end{table}

The held-out planner chose support balancing 15,501 times, direct plug-in
planning 7,354 times, equal-count partitions 1,708 times, and random-order
partitions 19,687 times. Population-oracle, strong, weak, and uninformative
score strata had mean effects over support balancing of 0.12166, 0.11438, 0.00717,
and 0.00041. Their medians were 0.08816, 0.08513, zero, and zero.

\section{Independent-Model Results}

\begin{figure}[t]
\centering
\begin{subfigure}[t]{0.48\textwidth}
\centering
\includegraphics[width=\textwidth]{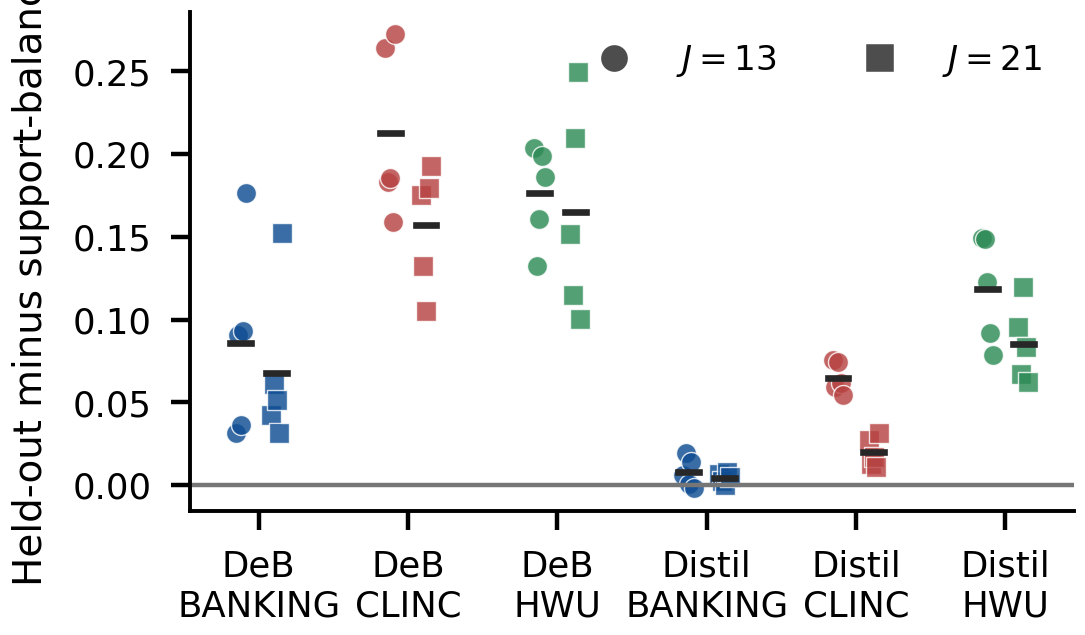}
\caption{All five model effects.}
\label{fig:real-c}
\end{subfigure}
\hfill
\begin{subfigure}[t]{0.48\textwidth}
\centering
\includegraphics[width=\textwidth]{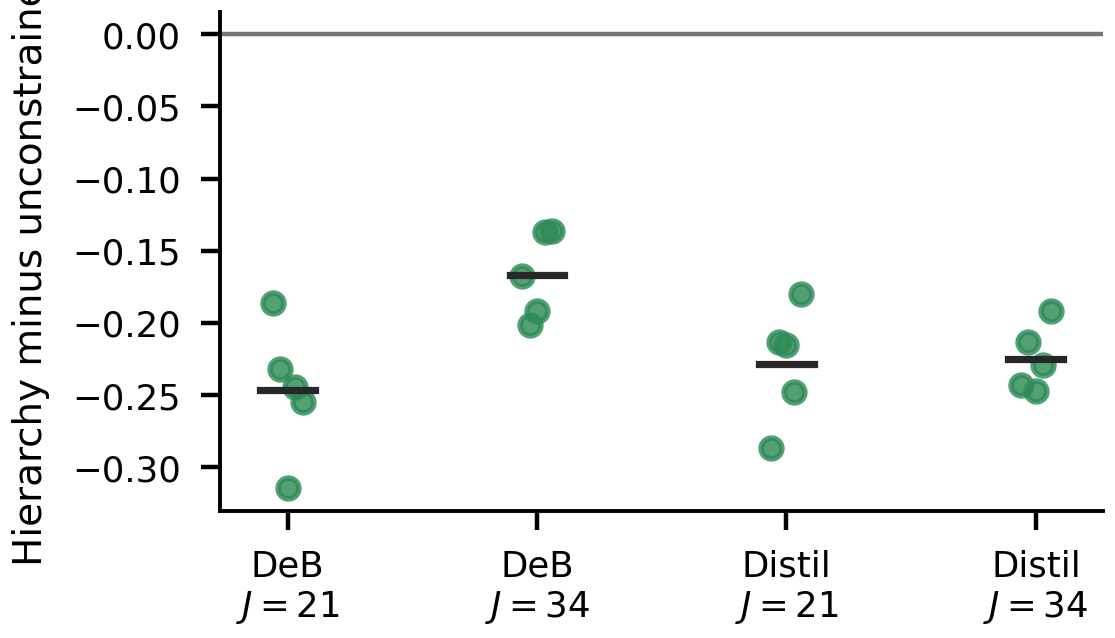}
\caption{Semantic constraints cost coverage.}
\label{fig:real-d}
\end{subfigure}
\caption{Per-model intent effects and the coverage cost of the HWU semantic
constraint.}
\label{fig:real-checks}
\end{figure}

Table~\ref{tab:real-effects} reports every primary model-level effect. Each
value averages 30 nested resplits within one trained model. These five
independent observations determine direction, range, and leave-one-model-out
summaries.

\begin{table}[h]
\centering
\caption{Held-out planner minus support-balanced IID coverage.}
\label{tab:real-effects}
\begin{tabularx}{\textwidth}{@{}lllX@{}}
\toprule
Architecture & Dataset & $J$ & Five model-level effects \\
\midrule
DeBERTa & BANKING & 13 & 0.0316, 0.0906, 0.0362, 0.0933, 0.1767 \\
        &         & 21 & 0.0426, 0.0613, 0.0513, 0.0316, 0.1525 \\
DeBERTa & CLINC   & 13 & 0.2639, 0.1833, 0.1857, 0.1590, 0.2723 \\
        &         & 21 & 0.1753, 0.1325, 0.1055, 0.1795, 0.1926 \\
DeBERTa & HWU     & 13 & 0.2034, 0.1325, 0.1607, 0.1987, 0.1863 \\
        &         & 21 & 0.1516, 0.1148, 0.2098, 0.2494, 0.1006 \\
DistilRoBERTa & BANKING & 13 & 0.0060, 0.0195, 0.0005, 0.0143, $-0.0018$ \\
        &         & 21 & 0.0067, 0.0028, 0.0003, 0.0077, 0.0051 \\
DistilRoBERTa & CLINC & 13 & 0.0755, 0.0595, 0.0743, 0.0614, 0.0543 \\
        &       & 21 & 0.0271, 0.0130, 0.0172, 0.0108, 0.0313 \\
DistilRoBERTa & HWU & 13 & 0.1492, 0.1485, 0.1226, 0.0921, 0.0788 \\
        &     & 21 & 0.0954, 0.0671, 0.1194, 0.0833, 0.0624 \\
\bottomrule
\end{tabularx}
\end{table}

The two architectures span a wide range of accuracy and score quality
(Appendix~\ref{app:score}), so they broaden the evaluated regime rather than
serving as matched-accuracy replacements for each other. The official test
trace is carried alongside every frontier but labeled as a shifted mixture
throughout. No shifted value contributes to the primary table.

\section{Controlled-Shift Details}
\label{app:shift}

\begin{figure}[h]
\centering
\begin{subfigure}[t]{0.48\textwidth}
\centering
\includegraphics[width=\textwidth]{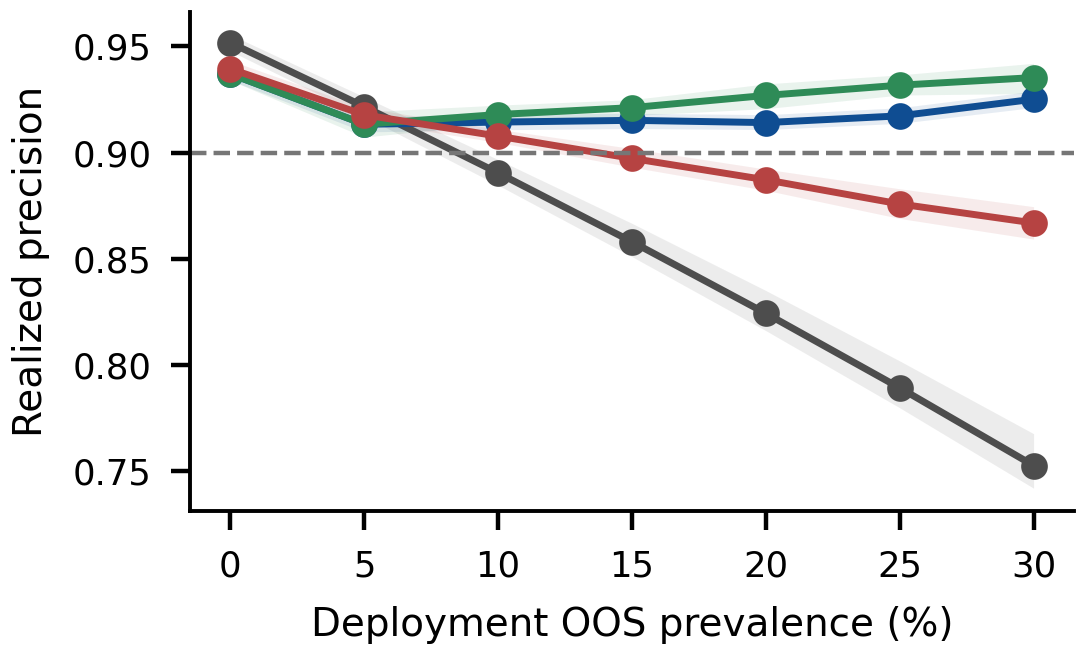}
\caption{Label-prior shift.}
\label{fig:shift-a}
\end{subfigure}
\hfill
\begin{subfigure}[t]{0.48\textwidth}
\centering
\includegraphics[width=\textwidth]{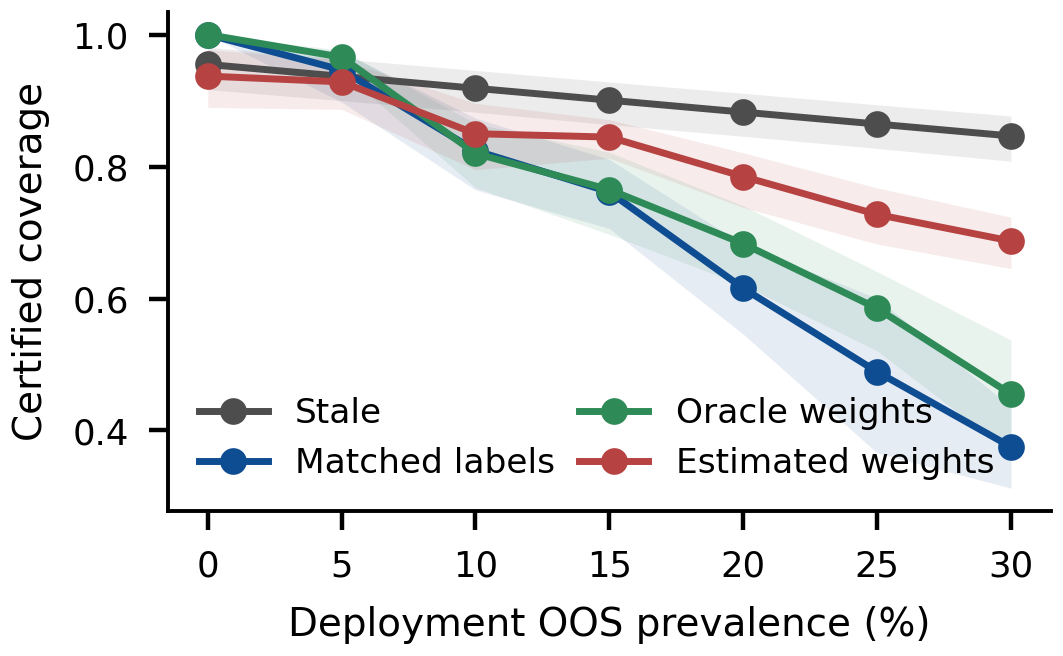}
\caption{Effective-support cost.}
\label{fig:shift-b}
\end{subfigure}

\begin{subfigure}[t]{0.48\textwidth}
\centering
\includegraphics[width=\textwidth]{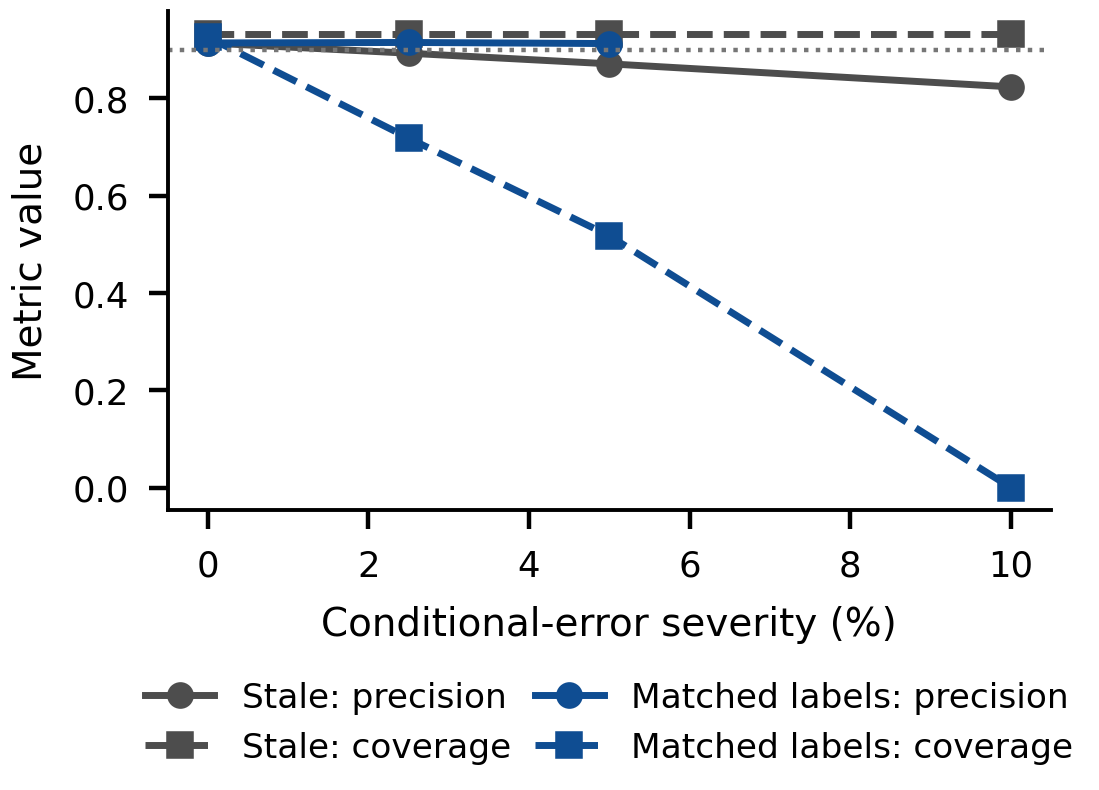}
\caption{Outcome drift requires new labels.}
\label{fig:shift-c}
\end{subfigure}
\hfill
\begin{subfigure}[t]{0.48\textwidth}
\centering
\includegraphics[width=\textwidth]{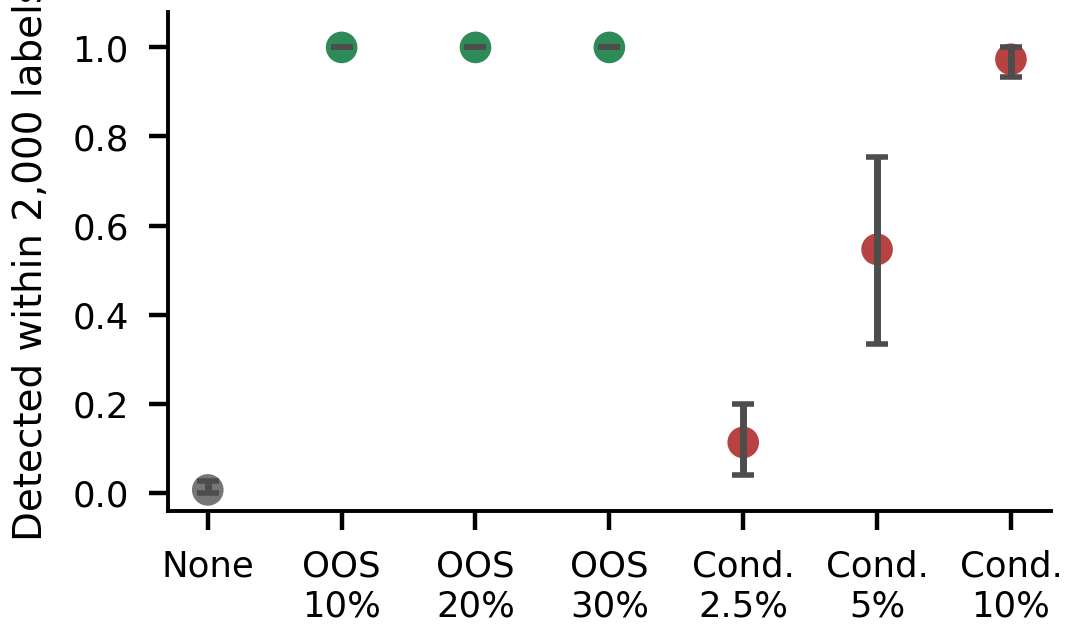}
\caption{Monitoring is delayed.}
\label{fig:shift-d}
\end{subfigure}
\caption{Controlled CLINC shift: label-prior repair and its cost, conditional
drift, and delayed monitoring.}
\label{fig:shift}
\end{figure}

Each of five CLINC models supplied 30 nested repetitions, drawn so that the
calibration and deployment pools remained disjoint within out-of-scope status.
Eleven frozen scenarios span the two kinds of shift. Some vary out-of-scope
prevalence, producing a pure label-prior shift; others convert a nested
fraction of high-confidence correct outcomes into errors, producing a
conditional-outcome shift. Against the first kind we compare three repair
strategies: exact rejection sampling from the known label-prior ratios, a
domain-classifier arm that estimates those ratios instead, and matched
recalibration on new deployment labels. A Bonferroni monitor watches two
signals at 40 checkpoints and labels 5\% of traffic. Throughout, precision is
conditioned on the serving repetitions, while coverage and feasibility retain
the abstain-all ones.

The headline results quantify each arm. Under 30\% out-of-scope prevalence,
oracle weighting reached 0.935 precision at 0.455 coverage, while estimated
weighting traded precision down to 0.867 for higher coverage of 0.687. Under
5\% conditional severity, matched new labels reached 0.913 precision at 0.517
coverage. By 10\% severity, no repetition certified at all.

\section{Extended Cross-Domain Results}
\label{app:multidomain-curves}
\begin{table}[t]
\centering
\caption{Certified coverage at every reporting granularity $J$ for the four
cross-domain gates, at three safety targets, over ten calibration resplits of one
frozen predictor per domain ($J{=}1$ is the global gate).}
\label{tab:md-curves}
\small
\begin{tabular}{@{}llrrr@{}}
\toprule
Domain & $J$ & $\tau{=}0.70$ & $\tau{=}0.80$ & $\tau{=}0.90$ \\
\midrule
Tool-call gating        & 1 & 1.00 & 0.79 & 0.17 \\
($K{=}4$, acc.\ 0.83)   & 2 & 1.00 & 0.67 & 0.17 \\
                        & 3 & 0.93 & 0.49 & 0.14 \\
                        & 4 & 0.84 & 0.41 & 0.09 \\
\midrule
Content moderation      & 1 & 1.00 & 1.00 & 0.97 \\
($K{=}5$, acc.\ 0.89)   & 2 & 1.00 & 1.00 & 0.95 \\
                        & 3 & 1.00 & 0.98 & 0.91 \\
                        & 5 & 1.00 & 0.91 & 0.85 \\
\midrule
Lesion classification   & 1 & 1.00 & 1.00 & 0.52 \\
($K{=}7$, acc.\ 0.86)   & 2 & 0.96 & 0.99 & 0.49 \\
                        & 3 & 0.95 & 0.98 & 0.55 \\
                        & 5 & 0.90 & 0.87 & 0.58 \\
                        & 7 & 0.81 & 0.72 & 0.46 \\
\midrule
Recommendation          & 1 & 1.00 & 0.65 & 0.32 \\
($K{=}18$, acc.\ 0.76)  & 2 & 1.00 & 0.79 & 0.27 \\
                        & 3 & 0.99 & 0.79 & 0.23 \\
                        & 5 & 0.98 & 0.76 & 0.22 \\
                        & 8 & 0.96 & 0.59 & 0.27 \\
                        & 13 & 0.91 & 0.45 & 0.17 \\
                        & 18 & 0.86 & 0.47 & 0.04 \\
\bottomrule
\end{tabular}
\end{table}

Table~\ref{tab:multidomain} reports the full safety-availability frontier for
the four cross-domain gates. For each safety target it gives certified coverage
at both the global unit and the finest unit, so a single row shows both edges
of the frontier: the cost of the target itself, and the additional cost of
splitting the gate into fine-grained units. The endpoint pairs illustrate the
price of higher targets and finer reporting, although intermediate coverage
need not decrease with every additional unit. Moderation loses almost nothing
until the target reaches 0.90. The tool-call gate has 0.79 global coverage at
0.80 and only 0.17 at 0.90. Recommendation, the least accurate model, pays the
steepest price, retaining only 0.04 of finely reported traffic at a 0.90
target. The final column reports a separate six-resplit held-out comparison at
one intermediate granularity and a domain-specific target. The gain is
substantial where the gate splits into several still-supportable units
(moderation, tool-calling) and near zero at the other two studied operating
points.

\begin{table}[h]
\centering
\caption{Cross-domain frontiers. Each pair is certified coverage at the global
and finest ($J{=}K$) units over ten resplits of one frozen model; the last column
is a separate six-resplit held-out gain, and Acc.\ is the per-task success
fraction rather than a common accuracy.}
\label{tab:multidomain}
\small
\begin{tabularx}{\textwidth}{@{}lXcccccc@{}}
\toprule
Domain & Model & $K$ & Acc. & $\tau{=}0.70$ & $\tau{=}0.80$ & $\tau{=}0.90$ & Held-out \\
\midrule
Tool-call gating & Qwen2.5-14B & 4 & 0.83 & 1.00/0.84 & 0.79/0.41 & 0.17/0.09 & $+0.051$ \\
Content moderation & logistic regression & 5 & 0.89 & 1.00/1.00 & 1.00/0.91 & 0.97/0.85 & $+0.176$ \\
Lesion classification & fine-tuned ResNet-18 & 7 & 0.86 & 1.00/0.81 & 1.00/0.72 & 0.52/0.46 & $+0.004$ \\
Recommendation & matrix factorization & 18 & 0.76 & 1.00/0.86 & 0.65/0.47 & 0.32/0.04 & $+0.008$ \\
\bottomrule
\end{tabularx}
\end{table}

Two boundaries of these studies are worth restating here. Each domain uses a
single frozen model, so the rows report how the trade-off behaves for that
particular model, not an effect size that would survive retraining. The
tool-call gate covers single-turn function selection, which speaks to when an
agent's individual calls can be certified but not to multi-turn or long-horizon
trajectories. Within those bounds, the table describes the same statistical
procedure applied to different success events and candidate populations
(Table~\ref{tab:contracts}).

\section{Exact Coverage and Additional Planner Comparisons}
\label{app:exact-checks}

The main text treats $U_J$ as an approximate objective and reports several
additional baselines. This appendix makes the approximation exact on problems
small enough that every state can be enumerated, so the error can be read off
with no Monte Carlo noise. It also specifies the additional planners and
controls that those comparisons use.

\subsection{Random selected support}
Condition on planning so that the score, reporting map, and starting thresholds
are frozen. For a segment $S$ at threshold $t$, selected certification support
is $M\sim\mathrm{Bin}(N,q_S(t))$. Conditional on $M=m$, the errors follow
$\mathrm{Bin}(m,r_S(t))$. Thus the exact expected population
coverage of this starting candidate is
\begin{equation}
v_{\mathrm{start}}(S,t;J)=q_S(t)\sum_{m=0}^{N}
\Prob(M=m)\,\pi(m,r_S(t);1-\tau,\delta/J),
\label{eq:integrated}
\end{equation}
with $\pi(0,r;\alpha,\gamma)=0$ and zero contribution when $q_S(t)=0$.
Population coverage is measured on independent traffic, not on the
certification sample. At $N=58$, $q=0.5$, $r=0$, $\tau=0.9$, $\delta=0.05$,
and $J=1$, certification requires at least 29 selected examples. The projected
score in Equation~\eqref{eq:segment} is therefore 0.500, whereas
Equation~\eqref{eq:integrated} is $0.5\Prob(M\geq29)=0.2761$.
Replacing random support by its mean is not a conservative approximation.

\subsection{The deployed threshold sequence}
Let $t_0>\cdots>t_L$ be the frozen grid starting at the planned threshold,
$q_\ell=q_S(t_\ell)$, and $B_\ell$ the event that test $\ell$ passes.
Stopping at the first failure gives exact expected coverage
\begin{equation}
v_{\mathrm{seq}}=
q_0\Prob(B_0)+\sum_{\ell=1}^{L}
(q_\ell-q_{\ell-1})\Prob(B_0\cap\cdots\cap B_\ell).
\label{eq:sequence}
\end{equation}
For each certification realization, coverage is the sum of these nonnegative
traffic increments whose preceding tests all passed. Taking expectations proves
the formula. Consequently $v_{\mathrm{seq}}\geq v_{\mathrm{start}}(S,t_0;J)$
for the same frozen policy. This population inequality does not make a
planning-set estimate a lower confidence bound, nor does it imply that the
starting objective maximizes deployed coverage. Refitting a score changes the
policy to which the calculation applies.

Across disjoint segments, expected coverage is additive by linearity of
expectation; independence of unit counts is unnecessary. Equation~\eqref{eq:dp}
therefore also optimizes exact starting or sequence values when supplied those
segment values, for a fixed order and $J$. It does not search noncontiguous
partitions. The reported large experiments retain their original projected
objective; the following exact study checks what that approximation misses.

\subsection{Independent exact reference}
The 36 cells cross $N=20,30,40$, $\tau=0.7,0.8$, balanced/skewed traffic,
and $J=1,2,3$, at $\delta=0.2$. Four groups have two score levels.
For each nonempty group subset, multinomial counts enumerate high-score
correct/error, low-score correct/error, and outside examples. Both possible
starting thresholds and all partitions are evaluated. The exact count-state
sizes are 10,626, 46,376, and 135,751. The independent multinomial calculation
and binomial integration agree within $4.9\times10^{-15}$.

\begin{table}[t]
\centering\small
\caption{Exact objective comparison over 36 finite problems. Regret is evaluated
using exact expected deployed coverage, with no Monte Carlo error.}
\begin{tabular}{@{}lrrr@{}}
\toprule
Quantity & Mean & Maximum & Positive cells\\
\midrule
Integrated-start improvement over projected-start & 0.0077 & 0.1440 & 8\\
Integrated-start regret to contiguous sequence optimum & 0.0079 & 0.2266 & 6\\
Contiguous sequence regret to unrestricted optimum & 0.0189 & 0.1368 & 10\\
\bottomrule
\end{tabular}
\end{table}

\subsection{Planning accuracy and meaningful service}
\label{app:planning-diagnostics}

\subsubsection{What an accurate planning score can establish}
For a finite pool of frozen candidates, let $S_c$ denote the exact expected
starting coverage, $D_c$ the expected deployed coverage under the downward
sequence, and $\widehat S_c$ the planning score. If $\widehat c$ maximizes that
score, define $\epsilon=\max_c|\widehat S_c-S_c|$ and
$\Delta=\max_c(D_c-S_c)$. Equation~\eqref{eq:sequence} gives $\Delta\geq0$,
which yields the elementary diagnostic bound
\begin{equation}
\max_c D_c-D_{\widehat c}\leq 2\epsilon+\Delta.
\label{eq:planning-regret}
\end{equation}
Indeed, for a deployed-coverage maximizer $c^*$, the left-hand side is at most
$S_{c^*}-S_{\widehat c}+\Delta$, which in turn is at most
$\widehat S_{c^*}-\widehat S_{\widehat c}+2\epsilon+\Delta$.
The score difference is nonpositive. Accurate starting scores therefore control
only part of the loss: the threshold sequence can change which candidate is
best. This is an elementary finite-pool inequality, not a new confidence bound
or a guarantee against partitions absent from the pool. Its unknown population
terms are diagnostics, not quantities available to the deployed planner.

\subsubsection{Intermediate-size integration comparison}
Before execution, we froze a sixteen-setting design crossing $K=8,16$, total
label budgets of 400 and 1{,}200, balanced versus Zipf traffic, and strong
versus weak scores, using the existing heterogeneous-error generator. Targets
are $\tau=0.9$ and $\delta=0.1$, with five independent plans per setting.
Twenty percent of labels go to planning and the remainder to certification.
Both $J=3$ and $J=5$ are evaluated on each plan with 100 shared, nested
certification draws. Both planners use the same empirical traffic frequencies,
Jeffreys error estimates, reliability order, and twenty thresholds
$0,0.05,\ldots,0.95$. The only difference is the segment objective:
Equation~\eqref{eq:segment} for projection, or integration over every support
count in Equation~\eqref{eq:integrated}. Both planners use the same dynamic
program and stop at the first failed certification test. All failures and
abstain-all outcomes are retained.

Mean deployed coverage is 0.2699 for projection, 0.2734 for integration, and
0.2275 for the support-balanced reference. Integration changes a map or starting
threshold in 107 of 160 plan-by-$J$ comparisons. Averaging the two
granularities within each of eighty independent plans yields sixteen wins,
seventeen losses, and forty-seven ties, with effects ranging from $-0.0984$ to
$+0.1841$. The standard error of 0.0035 uses within-setting variation across
the five plans; it measures simulation error around this fixed-design mean, not
robustness on unseen tasks. Table~\ref{tab:integration-conditions} retains all
32 condition means. Integration remains a plug-in calculation with estimated
rates, not a truth-informed oracle or an optimizer of the full deployed
sequence.

On one CPU with single-threaded BLAS, mean planning times are 0.0042 and
0.1651 seconds for projection and integration, respectively. Timing includes
segment scoring and the dynamic program, but excludes shared data generation,
group ordering, cutoff-cache preparation, and certification. Those shared costs
and all inputs are saved separately; method timing order alternates by plan.
The complete execution, including the saved-result audit below, took 46.45 seconds.
The ratio is implementation- and machine-dependent, not an asymptotic claim.

\subsubsection{Score errors and condition dependence}
An exploratory audit of the saved matched-pool study compares each winning
candidate's estimated score with its true integrated starting coverage. Mean
signed overestimation is 0.3429 for construction-data selection and 0.0469 for
held-out selection. Mean regret to the best integrated starting coverage within
the same pool falls from 0.0723 to 0.0338. These observations support the
claim that score assessment improves when conducted on separate data, but they
do not isolate pure selection bias: score projection, estimated rates, and
selection among different frozen policies all enter the comparison.
Table~\ref{tab:selection-strata} includes every factor stratum, and the
supplement retains all 64 condition effects and their paired plan-level
outcomes. These post hoc summaries do not select a preferred regime or support
multiple unadjusted significance claims.

\subsubsection{Named-scenario service}
For HWU, we also replay the saved certificates against the eighteen declared
scenario boundaries. Every candidate preserves those boundaries. A scenario is
counted once if any of its units passes, regardless of how many units it
contains. Table~\ref{tab:scenario-service} pairs this count with traffic
coverage rather than assigning an invented exchange rate between the two. The
direct constrained planner has higher means on both measures than full selection
in each of four architecture-by-$J$ settings. This aggregate comparison does
not constitute simultaneous confidence of dominance for individual scenarios.
Coverage and scenario service can disagree against other baselines: on DeBERTa
at $J=21$, support balancing serves less traffic but reaches 5.46 scenarios
against the direct planner's 5.32. No result here establishes that a
stakeholder would accept the unserved intents or these trade-offs. The audit
reuses ten existing trained models and their nested resplits; it is not
independent application confirmation.

\subsubsection{Matched-pool mechanism controls}
To ensure that every selector chooses from an identical slate, each synthetic
construction role produces the same eight maps; the selectors differ only in how
they choose among them. The uniform control averages coverage over the available
candidates, yielding the expected coverage of a uniform random choice.
Construction and held-out selection apply the same projected score and differ
only in which data role supplies the rates. The empirical-traffic control sums
selected frequencies for units whose Jeffreys error estimate is at most
$\alpha$, dropping the power term. The random-only selectors draw from the
first one, three, or five random candidates. Primary comparisons keep scores
and starts frozen, while refit comparisons update the starts from all planning
observations (the synthetic scores are fixed by the generator; real-data
refitting also updates the fitted correctness score).

Every condition retains 20 independent planning draws with 100 paired
certification draws per plan, and a vectorized first-failure implementation is
checked against the original scalar certifier. The source artifacts store
per-plan effects, their Monte Carlo standard errors, and per-condition win/loss
summaries and factor strata. The design draws fresh samples from declared
generator regimes, so it is prospective within those regimes but does not
establish generalization to an unknown distribution of tasks.

As a calibration check, the largest condition-mean false-certification rate
across the fifteen evaluated methods is 0.0305 against a nominal familywise
level of 0.10; this is descriptive rather than a proof of calibration over new
populations. The audit stores pointwise 95\% exact binomial intervals
conditional on each frozen plan, together with Monte Carlo standard errors
across the twenty independent plan rates. Uniform controls, which average
fractional outcomes, receive plan-level standard errors instead of binomial
intervals, and no interval treats the 2{,}000 nested trials as independent
planning replications.

\subsubsection{Candidate pool size}
If most of the benefit is the act of selecting among diverse candidates, two
questions follow: does a larger random pool keep helping, and do the structured
candidates add anything on top of it? We sweep the number of random-order
candidates through $R\in\{1,5,10,25,50,100\}$ on the synthetic cells, scoring
every candidate once on the selection split and certifying the selected one with
the same first-failure certifier as the main study. Held-out selection over
random candidates alone improves quickly and then saturates: mean certified
coverage rises from 0.258 at $R=1$ to 0.274 at $R=5$ and no further, reaching
only 0.274 at $R=100$. A larger random pool is therefore not the missing
ingredient. Adding the three structured candidates (support-balanced,
equal-count, and direct plug-in) to the $R$ random ones raises the mean by about
0.010 at every pool size, but the median improvement is zero throughout, so the
structured candidates help in a minority of cells rather than in the typical one.
Both observations agree with the matched-pool decomposition: the working
ingredient is disciplined selection over a handful of diverse groupings, not the
size of the pool or the structure of its members.

\subsubsection{Synthetic mechanism design}
The mechanism study crosses label budgets 800/3,200, 12/40 classes,
balanced/Zipf traffic, homogeneous/heterogeneous errors, and strong/weak scores
at $\tau=0.9$ and $\delta=0.1$, evaluating each of these 32 settings at $J=3,8$.
Planning receives 20\% of the labels, split equally between construction and
selection, and the remainder is reserved for certification. The 640 independent
planning draws are shared between the two values of $J$, giving 64 conditions of
20 plans each with 100 certification draws per plan; those nested certification
outcomes are not additional independent plans. This design was frozen before the
synthetic draws, whereas the real replay below reuses previously inspected traces
and is exploratory.

\subsubsection{Scenario-preserving comparison}
Every HWU candidate here preserves the 18 source scenarios. The direct candidate
orders groups by scenario and then by construction reliability, forbidding
cross-scenario segments, while the fixed baselines give each scenario at least
one unit and assign the remaining units to the scenario with the largest support
per current unit, subject to its class count (the equal-count baseline uses class
count in this allocation, and within each scenario the usual balancing rule
applies). Random maps permute the within-scenario order and reuse the same
support allocation. The construction is feasible at both $J=21$ and $J=34$, and
every saved map is checked for boundary preservation. Because the direct dynamic
program also optimizes how units are allocated across scenarios, its advantage
does not isolate within-scenario ordering from that allocation. All selectors
share the same construction score and candidate slate, and refitted comparisons
share the final planning score.

\begin{table}[t]
\centering\small
\caption{Scenario-preserving HWU replay: full held-out selector minus each comparator, as five per-model effects.}
\label{tab:constrained}
\begin{tabularx}{\textwidth}{@{}lllX@{}}
\toprule
Architecture & $J$ & Comparator & Five model effects\\
\midrule
deberta & 21 & Support & 0.0345, 0.0076, 0.0240, 0.0150, -0.0037\\
deberta & 21 & Random selection & 0.0361, -0.0043, -0.0009, 0.0101, -0.0181\\
deberta & 21 & Direct constrained & -0.0302, -0.0404, -0.0708, -0.0805, -0.0368\\
deberta & 34 & Support & 0.1466, 0.1350, 0.1682, 0.1638, 0.1157\\
deberta & 34 & Random selection & 0.1324, 0.1407, 0.1505, 0.1452, 0.1123\\
deberta & 34 & Direct constrained & -0.0098, -0.0276, -0.0470, -0.0298, -0.0469\\
distilroberta & 21 & Support & 0.0329, 0.0304, 0.0507, 0.0104, 0.0291\\
distilroberta & 21 & Random selection & 0.0197, 0.0700, 0.0284, 0.0075, 0.0322\\
distilroberta & 21 & Direct constrained & -0.0162, -0.0294, -0.0212, -0.0334, -0.0427\\
distilroberta & 34 & Support & 0.1393, 0.1576, 0.1419, 0.1338, 0.1639\\
distilroberta & 34 & Random selection & 0.1232, 0.1384, 0.1385, 0.1192, 0.1622\\
distilroberta & 34 & Direct constrained & -0.0313, -0.0352, -0.0318, -0.0292, -0.0060\\
\bottomrule
\end{tabularx}
\end{table}
\begin{table}[t]
\centering\small
\caption{Application replay on a shared eight-candidate pool: mean certified coverage over ten resplits, all methods refit.}
\label{tab:domain-replay}
\begin{tabular}{@{}lrrrrrr@{}}
\toprule
Domain & $\tau$ & $J$ & Support & Direct & Full selection & Random selection\\
\midrule
Function calls & 0.70 & 3 & 0.8464 & 0.9159 & 0.9013 & 0.9245\\
Proposed blocks & 0.90 & 3 & 0.8746 & 0.7312 & 0.7163 & 0.8786\\
Lesion images & 0.85 & 3 & 0.7239 & 0.7602 & 0.8726 & 0.7363\\
Proposed items & 0.85 & 5 & 0.3604 & 0.4475 & 0.3362 & 0.3077\\
\bottomrule
\end{tabular}
\end{table}
\begin{table}[t]
\centering\small
\caption{Per-condition random-support integration minus mean-support projection, with Monte Carlo standard errors across five plans.}
\label{tab:integration-conditions}
\begin{tabular}{@{}rrllrrr@{}}
\toprule
Groups & Labels & Traffic & Score & $J$ & Coverage difference & SE\\
\midrule
16 & 1200 & balanced & strong & 3 & 0.0066 & 0.0064\\
16 & 1200 & balanced & strong & 5 & -0.0444 & 0.0305\\
16 & 1200 & zipf & strong & 3 & 0.0000 & 0.0000\\
16 & 1200 & zipf & strong & 5 & 0.0000 & 0.0000\\
16 & 1200 & balanced & weak & 3 & 0.0007 & 0.0004\\
16 & 1200 & balanced & weak & 5 & -0.0001 & 0.0001\\
16 & 1200 & zipf & weak & 3 & 0.0000 & 0.0000\\
16 & 1200 & zipf & weak & 5 & -0.0009 & 0.0007\\
16 & 400 & balanced & strong & 3 & -0.0139 & 0.0139\\
16 & 400 & balanced & strong & 5 & 0.1191 & 0.0802\\
16 & 400 & zipf & strong & 3 & 0.0000 & 0.0000\\
16 & 400 & zipf & strong & 5 & 0.0156 & 0.0156\\
16 & 400 & balanced & weak & 3 & 0.0002 & 0.0002\\
16 & 400 & balanced & weak & 5 & 0.0002 & 0.0002\\
16 & 400 & zipf & weak & 3 & 0.0000 & 0.0000\\
16 & 400 & zipf & weak & 5 & 0.0000 & 0.0000\\
8 & 1200 & balanced & strong & 3 & 0.0083 & 0.0083\\
8 & 1200 & balanced & strong & 5 & 0.0015 & 0.0015\\
8 & 1200 & zipf & strong & 3 & -0.0158 & 0.0110\\
8 & 1200 & zipf & strong & 5 & 0.0148 & 0.0140\\
8 & 1200 & balanced & weak & 3 & -0.0017 & 0.0015\\
8 & 1200 & balanced & weak & 5 & -0.0001 & 0.0004\\
8 & 1200 & zipf & weak & 3 & -0.0001 & 0.0001\\
8 & 1200 & zipf & weak & 5 & 0.0000 & 0.0000\\
8 & 400 & balanced & strong & 3 & 0.0539 & 0.0445\\
8 & 400 & balanced & strong & 5 & 0.0092 & 0.0088\\
8 & 400 & zipf & strong & 3 & 0.0000 & 0.0000\\
8 & 400 & zipf & strong & 5 & -0.0402 & 0.0391\\
8 & 400 & balanced & weak & 3 & 0.0000 & 0.0000\\
8 & 400 & balanced & weak & 5 & 0.0000 & 0.0000\\
8 & 400 & zipf & weak & 3 & 0.0000 & 0.0000\\
8 & 400 & zipf & weak & 5 & 0.0000 & 0.0000\\
\bottomrule
\end{tabular}
\end{table}
\begin{table}[t]
\centering\small
\caption{Factor summaries of full-pool minus random-five held-out selection; ranges are over condition means, not confidence intervals.}
\label{tab:selection-strata}
\begin{tabular}{@{}llrrrr@{}}
\toprule
Factor & Value & Mean & Minimum & Maximum & Win/tie/loss\\
\midrule
Groups & 12 & 0.0054 & -0.0380 & 0.1202 & 19/2/11\\
Groups & 40 & -0.0000 & -0.0870 & 0.1344 & 10/2/20\\
Error variation & heterogeneous & -0.0034 & -0.0870 & 0.1098 & 11/1/20\\
Error variation & homogeneous & 0.0087 & -0.0852 & 0.1344 & 18/3/11\\
Labels & 3200 & -0.0057 & -0.0870 & 0.0473 & 12/1/19\\
Labels & 800 & 0.0110 & -0.0566 & 0.1344 & 17/3/12\\
Score & strong & 0.0054 & -0.0870 & 0.1344 & 18/0/14\\
Score & weak & -0.0000 & -0.0011 & 0.0021 & 11/4/17\\
Traffic & balanced & -0.0011 & -0.0870 & 0.1344 & 13/1/18\\
Traffic & zipf & 0.0064 & -0.0543 & 0.1202 & 16/3/13\\
\bottomrule
\end{tabular}
\end{table}
\begin{table}[t]
\centering\small
\caption{HWU service under fixed scenario boundaries: served traffic and the number of the eighteen scenarios with at least one passing unit, over ten resplits of five models.}
\label{tab:scenario-service}
\begin{tabular}{@{}lr lrr@{}}
\toprule
Architecture & $J$ & Planner & Coverage & Scenarios with service\\
\midrule
deberta & 21 & Support & 0.3252 & 5.46\\
deberta & 21 & Direct constrained & 0.3924 & 5.32\\
deberta & 21 & Full selection & 0.3406 & 5.26\\
deberta & 21 & Random selection & 0.3360 & 5.48\\
deberta & 34 & Support & 0.1742 & 3.52\\
deberta & 34 & Direct constrained & 0.3523 & 4.58\\
deberta & 34 & Full selection & 0.3201 & 4.38\\
deberta & 34 & Random selection & 0.1839 & 3.72\\
distilroberta & 21 & Support & 0.2246 & 3.88\\
distilroberta & 21 & Direct constrained & 0.2839 & 4.16\\
distilroberta & 21 & Full selection & 0.2553 & 4.04\\
distilroberta & 21 & Random selection & 0.2237 & 3.92\\
distilroberta & 34 & Support & 0.0861 & 1.94\\
distilroberta & 34 & Direct constrained & 0.2601 & 3.64\\
distilroberta & 34 & Full selection & 0.2334 & 3.46\\
distilroberta & 34 & Random selection & 0.0971 & 2.08\\
\bottomrule
\end{tabular}
\end{table}

\section{Familywise Budget Allocation}
\label{app:budget}

The deployed certifier spends an equal Bonferroni share $\delta/J$ of the
familywise budget on every unit. This is not the only validity-preserving choice:
any frozen allocation $\gamma_1,\ldots,\gamma_J$ with $\sum_j\gamma_j\le\delta$
certifies each unit $j$ at its own level $\gamma_j$, and a union bound over units
still gives familywise control at level $\delta$. This appendix asks how much
expected certified coverage the best such allocation can add over the equal
split, at the population level where the traffic share $q_j$ and true error
$r_j$ of each unit are known. The answer is an opportunity ceiling, not a
deployable gain: it uses the true errors and so bounds what any estimator of the
allocation could hope to recover, exactly as the population oracle bounds the
partition planner.

We sweep a factorial of $J\in\{3,5,8\}$ units, certification budgets
$N\in\{800,3200\}$, balanced or heavy-tailed (Zipf) traffic, homogeneous or
heterogeneous true errors, and targets $\tau\in\{0.80,0.90\}$ at $\delta=0.10$,
drawing 150 random unit configurations per cell. For each configuration we
compare the equal split against the coverage-maximizing allocation, computed
exactly by a $(\max,+)$ dynamic program over a fine budget grid; grid
discretization makes the reported gain a slight underestimate. Across the 7{,}200
configurations the optimal allocation adds a mean of 0.038 certified coverage
(median 0.028, maximum 0.288), exceeding 0.05 in about 30\% of configurations.
The gain grows exactly where the equal split is most wasteful
(Table~\ref{tab:budget}): with more units, a stricter target, and a scarcer
budget, where a few pressed units would otherwise consume their share without
certifying. Because this population opportunity is an order of magnitude larger
than the 0.0027 contributed by the structured partition candidate, we regard the
allocation of the testing budget, rather than the search over partitions, as the
more promising direction for closing the gap to the oracle.

The population study uses the true errors, so a fair question is whether a
deployable estimator can capture any of it. We repeat the sweep in a deployable
form: for each drawn truth we take a single planning sample, form Jeffreys error
estimates and frequency traffic estimates, allocate $\gamma_j$ by the same exact
dynamic program on those estimates alone, freeze it, and then evaluate the frozen
policy exactly at the true $(q_j, r_j)$ by integrating over random certification
support. The equal split is evaluated identically, and the paired difference is
the honest deployable gain. Across 2{,}880 truths the estimator improves realized
coverage by a mean of 0.011 (median 0.005), is positive on 68\% of truths, and
reaches a cell mean of 0.075 in the hardest regime (eight units, target 0.90,
scarce budget, heavy-tailed traffic); it turns slightly negative only in easy
regimes where the equal split is already near optimal and estimation noise
misallocates. The deployable gain is thus about a quarter of the population
ceiling, yet several times the structured-candidate contribution, and it is
realized without any change to the validity contract. A validity audit that
injects unsafe units ($r_j>\alpha$) across 3{,}800 configurations confirms this:
the allocated procedure's familywise false-certification probability averages
0.0002 and never exceeds 0.02, comfortably within the nominal $\delta=0.10$, as
the union bound guarantees. The estimator can also be improved by accounting for
uncertainty in the error rate. Replacing the Jeffreys plug-in with the
posterior-predictive availability, the exact average of the availability over the
Beta posterior of $r_j$ (a Beta-Binomial tail probability), raises the deployable
gain from 0.011 to 0.014 and its positive rate from 68\% to 74\%, recovering
close to two-fifths of the population ceiling; a purely pessimistic upper-quantile
rule, by contrast, hurts (mean 0.005), so it is the calibrated posterior average,
not conservatism, that helps.

\begin{table}[h]
\centering
\caption{Mean coverage gained by optimal familywise-budget reallocation over the
equal $\delta/J$ split, averaged over budgets, traffic shapes, and error
profiles.}
\label{tab:budget}
\small
\begin{tabular}{@{}lrr@{}}
\toprule
Units $J$ & $\tau{=}0.80$ & $\tau{=}0.90$ \\
\midrule
3 & 0.016 & 0.026 \\
5 & 0.027 & 0.047 \\
8 & 0.043 & 0.067 \\
\bottomrule
\end{tabular}
\end{table}

\subsection{Two ablations: the reliability order and the selection split}
Two design choices in the planner invite scrutiny, and we test both on the
synthetic cells with the first-failure certifier of the main study.

The dynamic program is optimal only over contiguous cuts of one frozen
reliability order, so a fair question is whether that order is a good structural
prior. It is not. Comparing the support-balanced partition built on the
reliability order against the same rule built on twenty random orders, the
reliability order beats the median random order only 33\% of the time and the
best of the twenty only 9\%, and its mean certified coverage is 0.249 against
0.282 for a random order. A random order is thus usually better, which is exactly
why the planner leans on random-order candidates and why selecting among diverse
groupings, rather than trusting the estimated order, is what carries the method.

The second choice is the single fifty-fifty split of the planning data into
construction and selection. Holding the candidate pool fixed and varying only how
much data scores it, selecting on all of the planning data, or on a five-fold
cross-fit of it, does not beat selecting on one held-out half: the median change
in realized coverage is zero and the mean is slightly negative, because scoring
on data that also built the candidates reintroduces exactly the optimism the
split was meant to remove. Cross-fitting is therefore not the missing ingredient;
the disjoint split is already doing its job, and the remaining gap to the oracle
lies in the objective and the budget, not in how the selection data is reused.

\section{Application Populations and Reproduction Details}
\label{app:application-details}

A shared trace format encodes binary success as label equality for
implementation convenience; this format does not make the underlying tasks
identical. BFCL uses the official abstract syntax tree (AST) checker for a frozen Qwen2.5-14B
single-turn trace \citep{patil2025bfcl}. Civil Comments \citep{borkan2019} and
MovieLens \citep{harper2015} use prefiltered candidates. DermaMNIST
\citep{yang2023medmnist} supplies a seven-class vision benchmark. After the
deterministic cap, BFCL has 753 evaluation records, Civil Comments 8{,}000,
DermaMNIST 1{,}003, and MovieLens 8{,}000. Certification and IID deployment are
disjoint roles sampled from those pools.

\begin{table}[t]
\centering\small
\caption{Application contracts. Coverage is the served fraction of the stated
candidate population, not necessarily the original dataset or deployment traffic.}
\label{tab:contracts}
\begin{tabularx}{\textwidth}{@{}lXXX@{}}
\toprule
Benchmark & Candidate population & Success event & Reporting unit\\
\midrule
BFCL & Existing single-turn function-call records & Official AST check passes
& Four recorded tool categories\\
Civil Comments & Comments predicted toxic, in a constructed class mixture
& Toxicity label at least 0.5, irrespective of subtype correctness
& Predicted policy subtype\\
DermaMNIST & Validation lesion images & Predicted lesion class is correct
& Predicted class\\
MovieLens & Rated pairs with predicted liking probability at least 0.5
& Observed rating at least four & First listed item genre\\
\bottomrule
\end{tabularx}
\end{table}

\subsection{Trace construction}
The moderation source retains all toxic comments and samples non-toxic comments
to produce a 300{,}000-row pool with 48.1\% toxic records. A 60/20/20 split
trains and evaluates a word-level term-frequency inverse-document-frequency
(TF-IDF) logistic regression, while a ComplementNB classifier and a
character-level TF-IDF logistic regression provide auxiliary
signals. Non-toxic predictions fall outside the gate. A toxic comment counts as
a justified block even when the predicted subtype is wrong, so this study does
not certify subtype attribution.

The MovieLens model is a binary matrix-factorization classifier with
64-dimensional embeddings, trained for eight epochs with batch size 8{,}192, Adam
learning rate 0.01, and dropout 0.2. An 80/10/10 random rated-pair split
separates training, evaluation, and test sets. Item-kNN and eight dropout passes
provide auxiliary scores. Because users and items can appear in multiple splits
and the candidate population consists of observed ratings, interpretation as
prospective recommendation is limited; no cold-start or user-independent
deployment guarantee is established.

DermaMNIST uses 224-pixel images and a pretrained ResNet-18 fine-tuned for five
epochs with AdamW, learning rate $10^{-4}$, batch size 64, and weight decay
0.01. A separately fine-tuned ViT-B/16 supplies auxiliary confidence, and eight
dropout passes supply disagreement scores. These constitute multiple model
computations even though there is only one base-predictor instance in the
reported frontier. Clinical triage decisions, patient outcomes, and
patient-level generalization are outside the scope of this experiment.

\section{Reproducibility and Artifact Integrity}
\label{app:release}

All reported results are tied to preserved experimental artifacts and a fully
pinned software environment, so that every figure, table, and headline number
can be regenerated deterministically from the recorded inputs and reconstructed
dependencies.

The supplementary material contains the experiment implementations, their
tests, the preregistration records, and machine-readable result tables,
including raw outputs for the exactness and selection studies, the four
application evaluations, and the model traces needed to replay them. An
automated check recomputes every registered numerical claim and generated table
from these canonical results and reports any discrepancy, establishing
traceability from each reported value back to the split and model that produced
it. The accompanying documentation details the exact steps for reconstructing
the environment and regenerating the figures and manuscript.

\end{document}